%% file: iclr2025.tex
\documentclass{article} 
\usepackage{iclr2025,times}

\input{math_commands.tex}

\usepackage{hyperref}
\usepackage{url}
\usepackage{graphicx}
\usepackage[ruled,noend]{algorithm2e}
\usepackage{algorithmic}
\usepackage{subcaption}
\usepackage{adjustbox}
\usepackage{multirow}
\usepackage{wrapfig}
\usepackage{float}

\title{Possibility for Proactive Anomaly Detection}

\author{Jinsung Jeon$^1$\thanks{This work was done when he was at Yonsei University.}, Jaehyeon Park$^2$, Sewon Park$^3$, Jeongwhan Choi$^4$, \\ \textbf{Minjung Kim}$^3$\textbf{,} \textbf{Noseong Park}$^2$\\
\textsuperscript{1}University of California, San Diego, \textsuperscript{2}KAIST, \textsuperscript{3}Samsung SDS, \textsuperscript{4}Yonsei University \\\
\texttt{jjsjjs0902@gmail.com} \\
}

\iclrfinalcopy 
\begin{document}

\maketitle

\begin{abstract}
Time-series anomaly detection, which detects errors and failures in a workflow, is one of the most important topics in real-world applications. The purpose of time-series anomaly detection is to reduce potential damages or losses. However, existing anomaly detection models detect anomalies through the error between the model output and the ground truth (observed) value, which makes them impractical. In this work, we present a \textit{proactive} approach for time-series anomaly detection based on a time-series forecasting model specialized for anomaly detection and a data-driven anomaly detection model. Our proactive approach establishes an anomaly threshold from training data with a data-driven anomaly detection model, and anomalies are subsequently detected by identifying predicted values that exceed the anomaly threshold. In addition, we extensively evaluated the model using four anomaly detection benchmarks and analyzed both predictable and unpredictable anomalies. We attached the source code as supplementary material.
\end{abstract}

\section{Introduction}


Time-series anomaly detection, which detects anomalous points caused by failures and errors, is one of the most important topics in the time-series task~\citep{munir2018deepant,he2019temporal,zhao2020multivariate,deng2021graph,chen2021learning}. In practice, various real-world applications reduce potential damage by determining the anomaly of events that have occurred. While previous anomaly detection models have shown good results in time-series anomaly detection, they are still limited in preventing anomalies in advance. However, since some anomalies appear with small precursors before they occur, these precursors can help predict the anomaly. 

In practice, many applications rely on reactive approaches to address anomalies after they have happened, which inherently limits their ability to prevent damages proactively. In contrast, a proactive approach aims to detect anomalies before they occur, enabling early intervention and damage mitigation. By identifying potential issues ahead of time, proactive anomaly detection can significantly reduce risks and damages compared to reactive methods, which focus on minimizing harm after anomalies have already happened. This proactive strategy offers a critical advantage in preventing or mitigating the impact of anomalies.

\begin{figure*}[t]
\centering
\includegraphics[width=\textwidth]{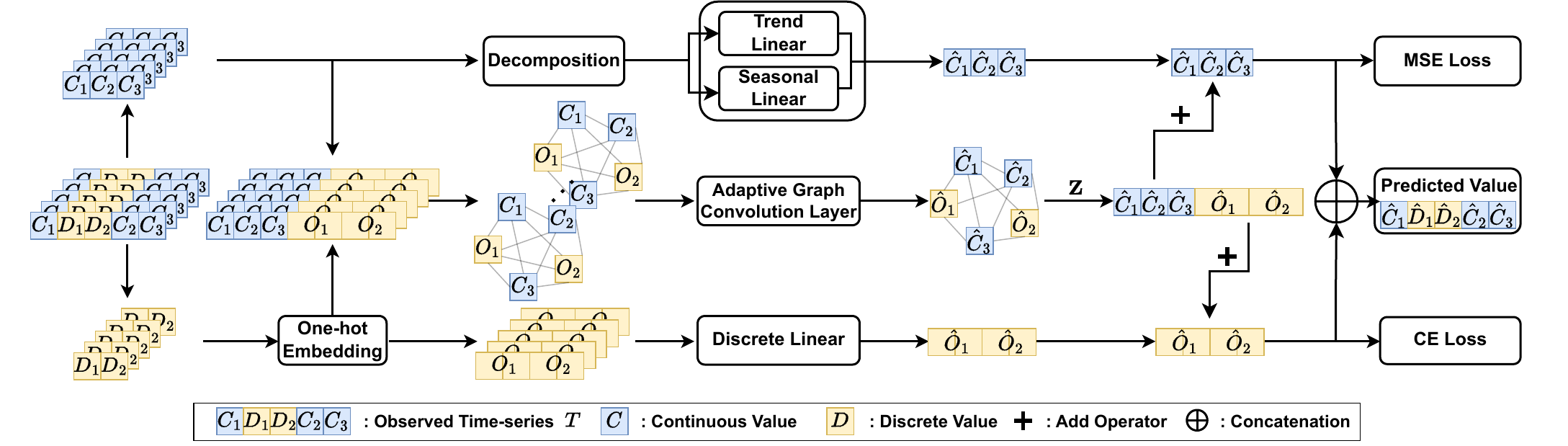} 
\caption{The architecture of our proposed time-series forecasting model specialized for anomaly detection in multivariate time-series data that considers both continuous and discrete values.} 
\label{fig:archi}
\vspace{-1em}
\end{figure*}

In this paper, we propose a \textit{proactive} approach for time-series anomaly detection, which integrates a time-series forecasting model specialized for anomaly detection (cf. Figure~\ref{fig:archi}). Our approach performs forecasting-based anomaly detection without relying on the ground truth of the testing data. It involves fitting a data-driven model using only training data, setting the extreme values of the distribution or boundary as a threshold, and then using the predicted values from our forecasting model as input to the trained data-driven model for anomaly detection.


\section{Related Work}

Data-driven anomaly detection relies entirely on the intrinsic properties of the data, such as density and distance. They learn the intrinsic properties of the training data and detect anomalies based on their respective criteria. Depending on the criteria, data-driven anomaly detection is generally divided into distribution-based methods and boundary-based methods. Distribution-based methods include the Gaussian mixture model (GMM) and empirical cumulative distribution-based outlier detection (ECOD). GMM assumes the training data follows a mixture of Gaussian distributions and uses log-likelihood values to detect outliers. ECOD~\citep{li2022ecod} estimates the joint cumulative distribution across all features and considers data points in the distribution's tail as anomalies.

Boundary-based methods encompass Support Vector Data Description (SVDD) and Deep Support Vector Data Description (DeepSVDD). SVDD~\citep{scholkopf1999support} obtains a hypersphere in high-dimensional feature space using kernel functions to encompass observations. DeepSVDD~\citep{ruff2018deep} employs a neural network to reduce high-dimensional data to a low-dimensional space before applying SVDD. Both approaches treat data points outside the learned hypersphere as anomalies.

\section{Proactive Anomaly Detection}


\subsection{Time-series Forecasting Model}\label{sec:model}
Proactive anomaly detection is based on a forecasting model for detecting anomalies in advance. Therefore, we design accurate predictive models suitable for anomaly detection datasets containing both continuous and discrete values. For this purpose, our model trains continuous and discrete features separately by each linear layer while utilizing a graph structure. 
Therefore, our model can consider the unique characteristics of both continuous and discrete features separately and learn the correlations between them simultaneously using the graph structure. This reduces errors in each feature and contributes to overall error reduction (See the Appendix for further details).

\begin{wrapfigure}{r}{5.5cm}
\vspace{-3em}
\centering
\includegraphics[width=0.40\textwidth]{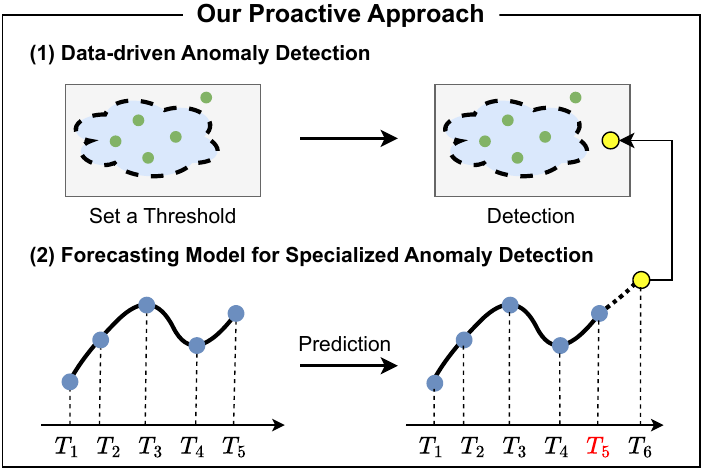} 
\caption{Overall process of our \textit{proactive} anomaly detection approach.} 
\label{fig:teaser}
\vspace{-2em}
\end{wrapfigure}

\subsection{How to Proactively Detect Anomaly}
Figure~\ref{fig:teaser} shows a process of proactive time-series anomaly detection. First, we train a data-driven model on the training data. Then, we extract a minimum value of training scores (which indicates a significant difference from the training data) from the trained model and set it as a threshold. Second, we extract the predicted value using testing data as input to the trained forecasting model. Finally, by using the predicted value as input to the trained data-driven model, it classifies the predicted value as normal if the score exceeds the threshold and an anomaly if the score is below the threshold.

\section{Experiments}
We conducted experiments on four real-world datasets. MSL, SMAP~\citep{hundman2018detecting}, SMD~\citep{su2019robust} and PSM~\citep{abdulaal2021practical} dataset. We use six time-series anomaly detection models and four time-series forecasting models to compare our forecasting model. 
To evaluate our detection model, we use F1-@K~\citep{kim2022towards}, F1-Composite~\citep{astha22evaluation}, and F1-Range~\citep{Wagner23timesead}, which are mainly used for anomaly detection (See the Appendix for further details).

\subsection{Proactive Anomaly Detection Performance} 
\begin{wraptable}{r}{8.3cm}
\vspace{-1.3em}
\caption{Results of anomaly detection experiment on the four benchmark datasets. Each number represents the ranking of the anomaly detection results across all models.}
    \label{tbl:main result}
\begin{center}
\vspace{-1em}
\begin{adjustbox}{width=0.6\textwidth}
\begin{tabular}{c|ccc|ccc|ccc|ccc}
    \hline
    Dataset & \multicolumn{3}{c|}{MSL} & \multicolumn{3}{c|}{SMAP} & \multicolumn{3}{c|}{SMD} & \multicolumn{3}{c}{PSM} \\ \hline
    Metric & F1-@K & F1-C & F1-R & F1-@K & F1-C & F1-R & F1-@K & F1-C & F1-R & F1-@K & F1-C & F1-R \\ \hline
    LSTM-P & 8 & 7 & 7 & 3 & 2 & 3 & 4 & 8 & 9 & 8 & 6 & 8 \\
    DeepAnT & 2 & \textbf{1} & 2 & 9 & 4 & 6 & \textbf{1} & 10 & 5 & 10 & 8 & 7 \\
    TCN-S2S-P & 3 & 4 & 4 & 2 & 6 & 3 & 8 & 9 & 4 & 8 & 10 & 11 \\
    MTAD-GAT & 6 & 3 & 2 & 5 & 3 & 2 & 11 & 6 & 11 & 7 & 9 & 9 \\
    GDN & 5 & 6 & 6 & 6 & 11 & 7 & 4 & 5 & 7 & 6 & 3 & 6 \\
    GTA & 4 & 5 & 5 & 11 & 6 & 9 & 9 & 11 & 5 & 11 & 10 & 10 \\ \hline
    NLinear & 10 & 9 & 10 & 6 & 9 & 8 & 3 & \textbf{1} & 3 & \textbf{1} & 4 & 2 \\
    DLinear & 10 & 11 & 10 & 6 & 9 & 11 & 6 & 3 & 2 & 4 & 4 & 4 \\
    TimesNet & 9 & 9 & 9 & 9 & \textbf{1} & 9 & 9 & 7 & 10 & 4 & 7 & 5 \\
    PatchTST & 7 & 8 & 8 & 3 & 4 & 5 & 2 & 3 & \textbf{1} & 2 & 2 & 3 \\ \hline
    Ours & \textbf{1} & \textbf{1} & \textbf{1} & \textbf{1} & 6 & \textbf{1} & 7 & \textbf{1} & 8 & 3 & \textbf{1} & \textbf{1} \\ \hline
    \end{tabular}
\end{adjustbox}
\end{center}
\vspace{-2em}
\end{wraptable}
Table~\ref{tbl:main result} summarizes the anomaly detection results based on rankings. We determined the ranking of each model for the data-driven models (GMM, ECOD, and DeepSVDD) by anomaly detection performance. After establishing the rankings for the three data-driven models, we calculated an integrated ranking by averaging across all metrics. We presented all results of our experiments in Appendix Table~\ref{tbl:results_msl} to~\ref{tbl:results_psm}.

Among the four benchmark datasets, all datasets except for PSM contain discrete values. Table~\ref{tbl:main result} shows that our model achieved the best performance on MSL and SMAP datasets with the highest ratio of discrete values. For datasets containing only continuous features, such as PSM, our model does not use prediction layers for discrete values. In this way, our forecasting model can be flexible even when there are only continuous features. 
As a result, it shows excellent performance in anomaly detection. 

\begin{wrapfigure}{r}{8.3cm}
\vspace{-1.5em}
\centering
    \includegraphics[width=0.5\textwidth]{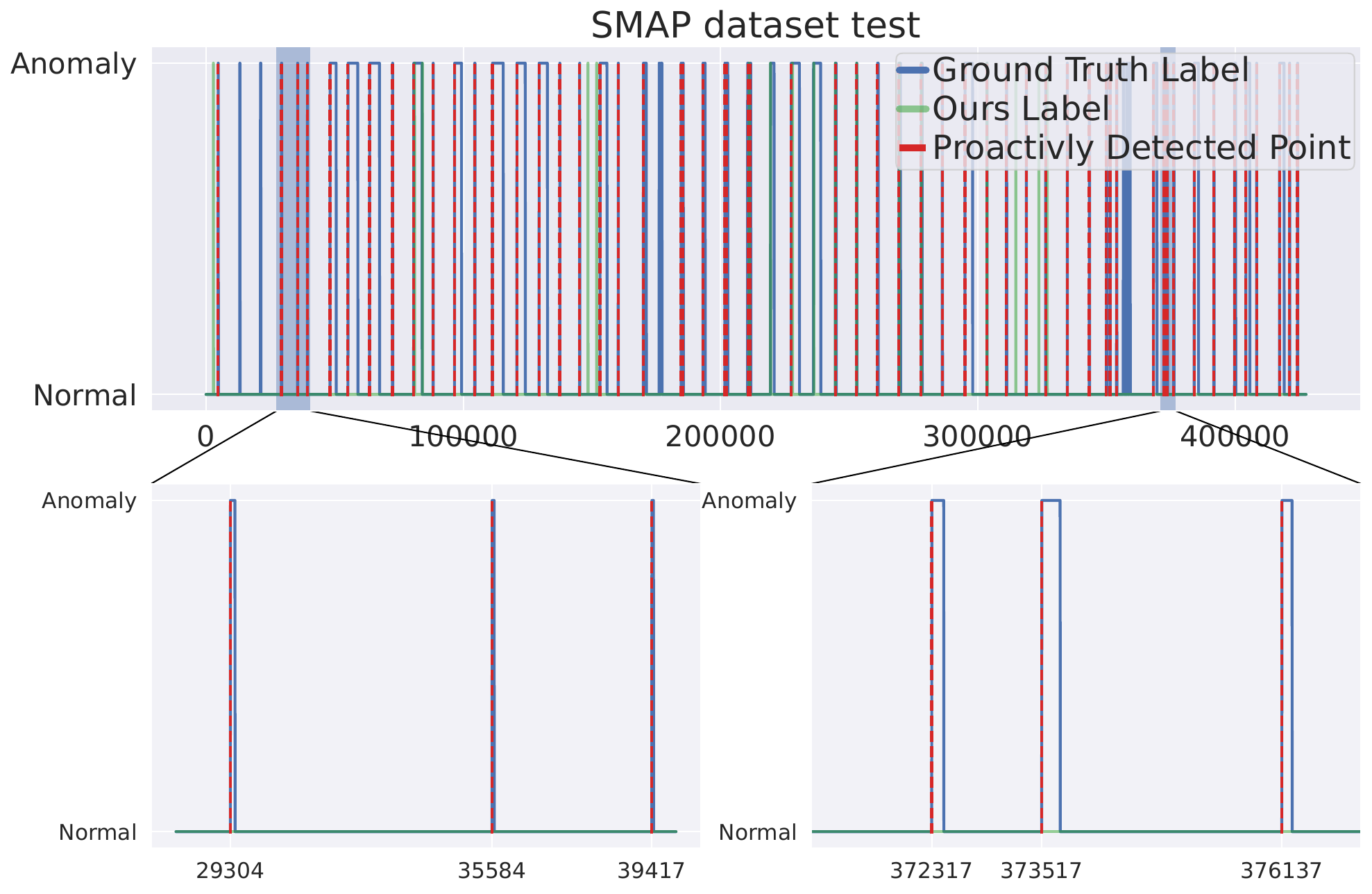}
      \caption{Visualization of proactive anomaly detection in SMAP dataset. The x-axis represents samples over time. Red line is the time point where proactive detection works.}
      \label{fig:casestudy3}
      \vspace{-1em}
\end{wrapfigure}

\subsection{Is Proactive Anomaly Detection Practical?}
This discussion question is the most important part of our proactive approach. Figure~\ref{fig:casestudy3} illustrates that our proactive approach identifies anomalies before abnormal situations occur in the SMAP dataset (Red lines). For example, an anomaly at timestamp 29304 is detected at timestamp 29303 before abnormal situations occur in the SMAP dataset (Figure~\ref{fig:casestudy3}, bottom left). Furthermore, in situations where abnormal situations persist, our proactive approach could minimize potential losses through early detection of the anomaly sample (Figure~\ref{fig:casestudy3}, bottom right).

\section{Forecastable Anomaly}
\label{sec:predictable}

Although successful, our proactive approach still has a drawback that makes it impractical. In real-world scenarios, some anomalies accompany small precursors before they happen (e.g., a volcano eruption), but there are also anomalies without precursors. One can consider that our method captures the anomalies with precursors. Therefore, we analyze to verify whether test anomalies can be forecasted by a time-series forecasting model. We rely on the Fourier transform for this analysis, which projects time-series onto orthogonal bases. Forecastable anomalies need to satisfy the following conditions. (i) The set of bases for test anomalies should be identical to that of training data. (ii) The coefficients of the bases for test anomalies should be within the range of those for training data. This predictability analysis using the convex hull of the bases has been conducted in various domains~\citep{yousefzadeh2021deep,yousefzadeh2020using}.

To analyze the test time-series with an anomaly in the frequency domain, we divide the time-series data into segments with a time length of 6, mirroring the forecasting process in our experiments, with five observations and one last anomaly. Then, each time-series segment $\{T_t \}^{6}_{t=1}$ can be represented as follows:
\begin{align}
    \begin{split}
    \{T_t \}^{\text{6}}_{t=1} =  a_1 f_1 + a_2f_2 + a_3 f_3 + a_4 f_4 ,
    \end{split}
    \label{eq:dft}
\end{align}
where each $f_i$ is a Fourier basis and $a_i$ is its corresponding coefficient calculated by the real-valued discrete Fourier transform (DFT). When the segment length is $N$, the DFT result is represented by $\frac{N}{2} + 1$ frequencies. In our proposed method, a time-series segment can be expressed using the Fourier bases in $F=\{f_1, f_2, f_3, f_4\}$ with the coefficients $A=(a_1, a_2, a_3, a_4)$. Different time-series segments can have different bases, although their set sizes are always 4. For convenience, we refer to the segmented training data and the segmented testing anomaly data as the training sample and the anomaly sample, respectively.

Given $N$ training samples and $M$ anomaly samples, we first calculate the superset of their bases and then verify whether these supersets are the same, i.e., $\cup_{i=1}^N F_i = \cup_{j=1}^M F_j$, where $F_i$ and $F_j$ are a set of bases for a training and a testing anomaly sample, respectively. Then we find that their supersets are identical in all our datasets used for experiments. Therefore, the first condition is satisfied. This analysis shows that all training and testing samples are in the same manifold, which is an amenable situation for a forecasting model to learn.

For the second condition, we also compare the basis coefficients of the training and anomaly samples. Given $N$ training samples, we first construct a convex polytope using the magnitude of $\{A_i\}_{i=1}^N$, where $A_i$ means the coefficients of the $i$-th training sample. This convex polytope is a region made by simple interpolations with the magnitude of $\{A_i\}_{i=1}^N$. The interpolation is one of the simplest types of the inferences that deep neural networks can do~\citep{raissi2019physics,tashiro2021csdi,harvey2022flexible,constantinescu2023interpolation}. Therefore, we check whether the coefficients' magnitudes of each testing anomaly sample in $\{A_j\}_{j=1}^M$, where $A_j$ means the coefficients of the $j$-th anomaly sample, lie inside or outside the convex region.

\begin{figure*}[t]
\centering
  \begin{subfigure}{\columnwidth}
    {{\includegraphics[width=0.5\textwidth]{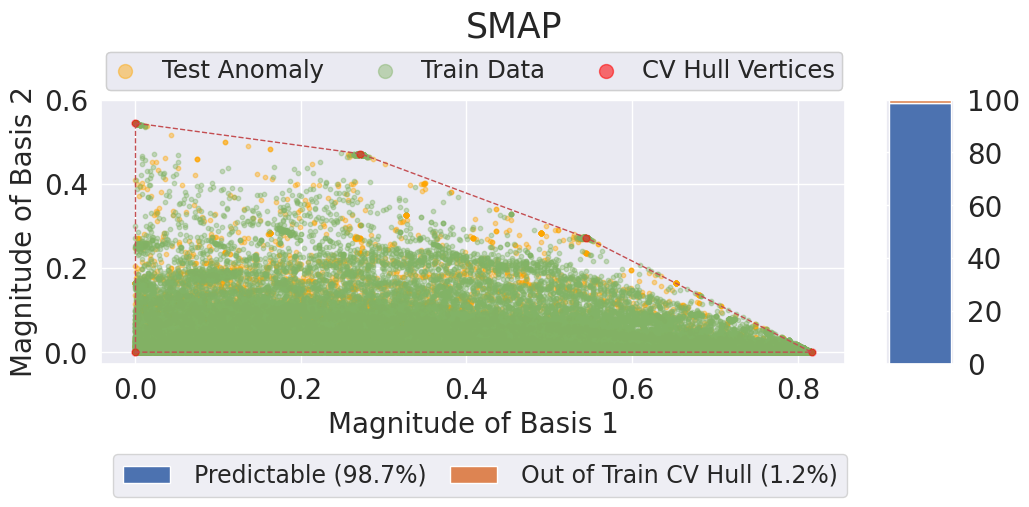}}}
    {{\includegraphics[width=0.5\textwidth]{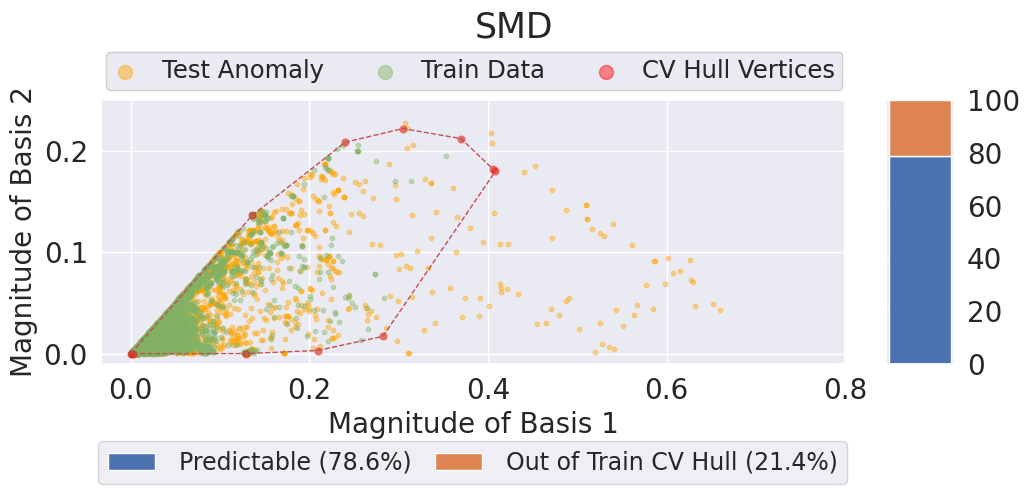}}}
  \end{subfigure}
\caption{Visualization of forecastable anomaly values. A convex hull (CV hull) is constructed using the magnitudes of the training data, which represent the absolute values of the coefficients. Anomalies are considered unforecastable if they have magnitudes outside the CV hull.}
\label{fig:predictable}
\vspace{-1.5em}
\end{figure*}

Figure~\ref{fig:predictable} shows two examples with SMAP and SMD. In each figure, a convex hull with two bases, a two-dimensional convex polytope, is shown since we cannot visualize higher than two in this paper --- however, our analysis is done in the complete convex polytope. In the case of the SMAP dataset in Figure~\ref{fig:predictable}, it shows that there are very few anomaly samples that are not included in the convex hull of the training data. In total, 1.2\% of the anomaly samples are not inside the convex polytope. On the other hand, for the SMD data, around 21\% of test time-series with an anomaly fall outside the convex polytope. We expect that it is difficult for our model to perform well on the SMD dataset compared to other datasets including SMAP.

\section{Conclusion}

Time-series anomaly detection is one of the important research topics in real-world applications. 
However, previous works have applied a $\textit{reactive}$ approach that cannot prevent anomalies in advance.
Our $\textit{proactive}$ anomaly detection is a forecasting-based approach specialized for anomaly detection in multivariate time-series data with continuous and discrete features. Thus, it is able to detect and prevent anomalies in advance by accurate forecasting.
Our experiments incorporating real-world datasets and various metrics support the efficacy of our specialized forecasting model in time-series anomaly detection.
In conclusion, we analyze both predictable and unpredictable anomalies in our proactive anomaly detection system, thereby leaving room for unpredictable anomalies. 

\subsubsection*{Acknowledgments}
This work was supported by the National Research Foundation of Korea (NRF) grant funded by the Korea government (MSIT) (No. RS-2024-00405991).

\bibliography{iclr2025}
\bibliographystyle{iclr2025}

\clearpage

\appendix



\setcounter{table}{0}
\renewcommand{\thetable}{\Alph{table}}
\setcounter{figure}{0}
\renewcommand{\thefigure}{\Alph{figure}}

\section{Forecasting Model for Detecting Anomalies}
In this section, we describe detailed a time-series forecasting model specialized for anomaly detection. For convenience of understanding, notations are defined as follows:
\begin{itemize}
    \item Observed time-series $\bm{T} \in \mathbb{R}^{(c+d) \times N} $ is denoted by a set of time points $\{T_1, T_2,\cdots, T_N \}$.
    \item The observed value at time $t$, $T_t \in \mathbb{R}^{c+d}$ contains $c$ continuous columns and $d$ discrete columns. It is expressed as $T_t = \{\mathbf{C}^t, \mathbf{D}^t \}$ at time $t$.
    \item $\mathbf{C}^t\in\mathbb{R}^{c}$ is denoted by a set of columns with a continuous value $\{C_1, C_2,\cdots, C_c \}^t$ at time $t$. 
    \item $\mathbf{D}^t\in\mathbb{R}^{d}$ is denoted by a set of columns with a discrete value $\{D_1, D_2,\cdots, D_d \}^t$ at time $t$.
\end{itemize}

\subsection{Extract Continuous \& Discrete Features.}\label{sec:features}
First, continuous and discrete features are separately extracted for prediction according to the data characteristics. Therefore, we divide continuous and discrete values for observed values $ \{T_t \}^N_{t=1} \in \mathbb{R}^{(c+d) \times N}$ as follows:
\begin{align}
    \begin{split}
    \{T_t \}^N_{t=1} &= \{\mathbf{C}^t, \mathbf{D}^{t} \}^N_{t=1}, \\
    &= \{  \mathbf{C}^t\}^N_{t=1}, \{\mathbf{D}^{t} \}^N_{t=1}.
    \end{split}
\end{align}

Then, in time-series prediction, the model considering trend and seasonality shows simple but excellent performance~\citep{zeng2023transformers}, so we decompose continuous values into trend and seasonality.
To extract the trend $\mathbf{C}_{\mathbf{T}}$ from the continuous values, we use the moving average method, and the seasonality $\mathbf{C}_{\mathbf{S}}$ is taken as the remaining value after subtracting the trend from the continuous values.
\begin{align}
    \begin{split}
    \{ \mathbf{C}^{t}_{\mathbf{T}}\}^N_{t=1} &= \text{AvgPool}(\text{Padding}(\{\mathbf{C}^t\}^N_{t=1})),\\
     \{\mathbf{C}^{t}_{\mathbf{S}}\}^N_{t=1} &= \{\mathbf{C}^t\}^N_{t=1} - \{\mathbf{C}^{t}_{\mathbf{T}}\}^N_{t=1},
    \end{split}
\end{align}
where AvgPool means average pooling and Padding means pre-padding with the first value and post-padding with the last value. Then, using each component as input to a linear layer, each hidden vector is extracted as follows:
\begin{align}
    \begin{split}
     \mathbf{\hat{C}}^{N+1}_\mathbf{T} &= \text{Linear}_{trend}(\{\mathbf{C}^{t}_{\mathbf{T}}\}^N_{t=1}),\\
     \mathbf{\hat{C}}^{N+1}_\mathbf{S}  &= \text{Linear}_{seasonality}(\{\mathbf{C}^{t}_{\mathbf{S}}\}^N_{t=1}),
    \end{split}
\end{align}
where $\text{Linear}_{trend}$ and $\text{Linear}_{seasonality}$ are linear layers that predict the next time point $N+1$ with $N$ observed time points. After the hidden vectors of each trend and seasonality are extracted, reconstruct continuous features by adding each hidden vector.
\begin{align}
    \begin{split}
        \mathbf{\hat{C}}^{N+1} = \mathbf{\hat{C}}^{N+1}_\mathbf{T} + \mathbf{\hat{C}}^{N+1}_\mathbf{S},
    \end{split}
\end{align}
where $\mathbf{\hat{C}}^{N+1} \in \mathbb{R}^{c}$ is denoted by a set of $\{\hat{C}_1, \hat{C}_2,\cdots, \hat{C}_c \}^{N+1}$. 

For discrete values, due to the difficulty in explaining trend and seasonality, they are converted into one-hot vectors and used as inputs for a linear layer to extract hidden vectors. The process is as follows:
\begin{align}
    \begin{split}
    \{\mathbf{O}^{t} \}^N_{t=1} &= \text{One-Hot}(\{\mathbf{D}^{t} \}^N_{t=1}), \\
    \mathbf{\hat{O}}^{N+1} &= \text{Linear}_{discrete}(\{\mathbf{O}^{t} \}^N_{t=1}),
    \end{split}
\end{align}
where One-Hot indicates a one-hot embedding, $\mathbf{O}^{t} \in \mathbb{R}^{d \times e}$ is a set of one-hot vectors of discrete columns, $e$ is an embedding dimension and 
$\mathbf{\hat{O}}^{N+1} \in \mathbb{R}^{d \times e}$ denotes a set of predicted one-hot vectors of discrete columns $\{\hat{O}_1, \hat{O}_2,\cdots, \hat{O}_d \}^{N+1}$. Then, $\hat{O}_t \in  \mathbb{R}^{e}$ is a predicted one-hot vector of discrete value. $\text{Linear}_{discrete}$ is a linear layer that predicts the next time point $N+1$ with $N$ observed time points.


\subsection{Adaptive Graph Convolution Layer.}\label{sec:agc}
Second, we utilize an adaptive graph convolution (AGC) layer. The lack of learning dependencies between features due to the separate handling of continuous and discrete values can be addressed by using an adaptive adjacency matrix. The adaptive adjacency matrix is formed by multiplying learnable embedding and is defined as follows:
\begin{align}
\mathbf{A} = \mathbf{I}+\text{softmax}(\text{ReLU}(\mathbf{E} \mathbf{E}^{\textsf{T}})),
\end{align} 
where $\mathbf{E}\in\mathbb{R}^{(c+d)\times b}$ is a trainable node-embedding matrix with embedding dimension $b$, $\mathbf{I}\in\mathbb{R}^{(c+d)\times (c+d)}$ is the identity matrix, and $\mathbf{A}\in\mathbb{R}^{(c+d)\times (c+d)}$ is the learned adjacent matrix of the graph representing the proximity between time-series features.

We combine the adaptive adjacency matrix and graph convolution network. 
To utilize AGC, the one-hot embedding of continuous features and discrete features are concatenated into $\mathbf{X}\in \mathbb{R}^{(c+d)\times (e\times N)}$.
$\mathbf{X}$ is reshaped from the $\{\tilde{\mathbf{C}}^t,\mathbf{O}^t \}^N_{t=1}\in\mathbb{R}^{(c+d)\times e \times N}$, where $\tilde{\mathbf{C}}^t=[\mathbf{C}^t,\mathbf{0}]\in \mathbb{R}^{(c+d)\times e}$ is zero-padded continuous features.
Let $\mathbf{H}\in \mathbb{R}^{(c+d)\times h}$ is the matrix of node features with hidden dimension $h$ transformed by $\text{Linear}_{input}$.
Then the AGC layer outputs $\mathbf{Z}$ through the following mapping:
\begin{align}
\mathbf{X} &= \text{reshape}(\{\tilde{\mathbf{C}}^t,\mathbf{O}^t \}^N_{t=1}), \label{eq:gcn1}\\
\mathbf{H} &= \text{Linear}_{input}(\mathbf{X}), \label{eq:gcn2}\\
\mathbf{Z} &= \sigma(\mathbf{A}\mathbf{H}\mathbf{E}\mathbf{W}), \label{eq:gcn3}
\end{align}where $\mathbf{W}\in \mathbb{R}^{b \times h \times e}$ is a trainable weight transformation matrix and $\sigma(\cdot)$ is the activation function.
$\mathbf{Z}\in \mathbb{R}^{(c+d)\times e}$ is the result of adaptive graph convolution and will be utilized in the prediction stage.

\subsection{Prediction.}
Finally, the dependencies between features are added to each extracted discrete and continuous hidden vector. Through this, it is possible to make predictions considering the dependence between the separately extracted hidden vectors. 
\begin{align}
    \begin{split}
     \mathbf{\hat{C}}^{N+1} &= \mathbf{\hat{C}}^{N+1} + \mathbf{Z}_{:c,1} ,\\
     \mathbf{\hat{O}}^{N+1} &= \mathbf{\hat{O}}^{N+1} + \mathbf{Z}_{-d:,:},
    \end{split}
\end{align}
where $\mathbf{Z}_{:c,1} \in \mathbb{R}^c$ is continuous columns and  $\mathbf{Z}_{-d:,:} \in \mathbb{R}^{d \times e}$ is discrete columns in $\mathbf{Z}$.

After that, continuous features are trained with the mean square error (MSE) loss as follows:
\begin{align}
    \begin{split}
     Loss_{\mathbf{C}} &= \frac{\sum_{i=1}^{c} (C^{N+1}_{i}-\hat{C}^{N+1}_{i})^{2}}{c},
    \end{split}
\end{align}
where $C^{N+1}_i$ is the $i$-th element of $\mathbf{C}^{N+1}$, $c$ is the number of continuous elements. For discrete features, we need to train the model with cross-entropy (CE) loss for each discrete feature as follows:
\begin{align}
    \begin{split}
     Loss_{\mathbf{D}} &= \frac{\sum_{i=1}^{d}\sum_{j=1}^{e}-O_{ij}^{N+1}log(\hat{O}_{ij}^{N+1})}{d},
    \end{split}
\end{align}
where $O_{ij}^{N+1}$ is the $j$-th one-hot vector element in $i$-th one hot vector $O_{i}^{N+1}$. $e$ is one-hot embedding dimension and $d$ is the number of discrete elements. The total loss of our prediction model is:
\begin{align}
    \begin{split}
     Loss_{\textbf{Total}} &= Loss_{\mathbf{C}} + Loss_{\mathbf{D}}.
    \end{split}
\end{align}

\subsection{Training Algorithm}
We present the training process of the proposed time-series forecasting model in Algorithm~\ref{alg:algorithm_prediction}. At each iteration, we first divide the continuous and discrete values in the observed time-series \textit{$\textbf{T}_{train}$}. Each of the continuous values and discrete values is passed through separate prediction models to predict $\mathbf{\hat{C}}^{N+1}$ and $\mathbf{\hat{O}}^{N+1}$. In order to consider the dependency between values regardless of continuous or discrete, $\mathbf{Z}$ is extracted by passing through the adaptive graph convolution (AGC) layer with the observed time-series \textbf{\textit{$T_{train}$}} as an input. Then add $\mathbf{Z}$ to $\mathbf{\hat{C}}^{N+1}$ and $\mathbf{\hat{O}}^{N+1}$. Finally, the prediction model is trained with the mean squared error loss for continuous values and the cross entropy loss for discrete values.

\begin{algorithm}[h]
\caption{How to train time-series forecasting model}
\label{alg:algorithm_prediction}
\textbf{Input}: training time-series data $\bm{T}_{\text{Train}}$, Iteration number of prediction model $K_{\text{Pred}}$  \\
\textbf{Parameter}: Prediction model $\boldsymbol{\theta}_{\text{Pred}}$  \\
\textbf{Output}: Prediction model $\boldsymbol{\theta}_{\text{Pred}}$  \\
\begin{algorithmic}[1] 
\STATE Initialize $\boldsymbol{\theta}_{\text{Pred}}$  
\STATE $k \gets 0 $ 
\WHILE{$k < K_{\text{Pred}}$}
\STATE $\{  \mathbf{C}^t\}^N_{t=1}, \{\mathbf{D}^{t} \}^N_{t=1} \gets \textbf{\textit{T}}_{\text{Train}}$ 
\STATE $\{  \mathbf{C}^{t}_{\mathbf{T}}\}^N_{t=1}, \{\mathbf{C}^{t}_{\mathbf{S}} \}^N_{t=1} \gets \text{Decomp} \{\mathbf{C}^t\}^N_{t=1}$ 
\STATE $ \mathbf{\hat{C}}^{N+1}_\mathbf{T} \gets \text{Linear}_{trend}(\{  \mathbf{C}^{t}_{\mathbf{T}}\}^N_{t=1})$
\STATE $\mathbf{\hat{C}}^{N+1}_\mathbf{S}\gets \text{Linear}_{seasonal}(\{ \mathbf{C}^{t}_{\mathbf{S}}\}^N_{t=1})$
\STATE $ \mathbf{\hat{C}}^{N+1} \gets \mathbf{\hat{C}}^{N+1}_\mathbf{T} + \mathbf{\hat{C}}^{N+1}_\mathbf{S}$
\STATE $ \mathbf{O}^{N+1}\gets \text{One-hot}(\{  \mathbf{D}^t\}^N_{t=1})$ 
\STATE $ \mathbf{\hat{O}}^{N+1}\gets \text{Linear}_{discrete}(\{  \mathbf{O}^t\}^N_{t=1})$ 
\STATE $\mathbf{Z} \gets$ AGC layer with Eq.~7 to Eq.~10
\STATE $\mathbf{\hat{C}}^{N+1} = \mathbf{\hat{C}}^{N+1} + \mathbf{Z}_{:c,1}$
\STATE $\mathbf{\hat{O}}^{N+1} = \mathbf{\hat{D}}^{N+1} + \mathbf{Z}_{-d:,:}$
\IF {Continuous value}
\STATE $Loss_{\mathbf{C}} = \frac{\sum_{i=1}^{c} (C^{N+1}_{i}-\hat{C}^{N+1}_{i})^{2}}{c}$
\ENDIF
\IF {Discrete value}
\STATE $Loss_{\mathbf{D}} = \frac{\sum_{i=1}^{d}\sum_{j=1}^{e}-O_{ij}^{N+1}log(\hat{O}_{ij}^{N+1})}{d}$
\ENDIF
\STATE $Loss_{\text{Total}} \gets Loss_{\mathbf{C}} + Loss_{\mathbf{D}}$
\STATE Update $\boldsymbol{\theta}_{\text{Pred}}$ with $Loss_{\text{Total}}$ 
\ENDWHILE
\STATE \textbf{return} Prediction model $\boldsymbol{\theta}_{\text{Pred}}$
\end{algorithmic}
\end{algorithm}

\section{Experimental Environments}

Our detailed software and hardware environments are as follows: \textsc{Ubuntu} 18.04 LTS, \textsc{Python} 3.9.12, \textsc{CUDA} 11.4, \textsc{NVIDIA} Driver 525.125.06 i9 CPU, and \textsc{GeForce RTX A5000} \& \textsc{A6000}.  

\subsection{Dataset}
We used four time-series datasets for our experiments. Summary of the datasets in Table~\ref{tbl:dataset} of the main manuscript.
\begin{itemize}
\item Mars Science Laboratory rover and Soil Moisture Active Passive satellite~\citep{hundman2018detecting} datasets are from NASA. Mars Science Laboratory (MSL) rover contains 55 features, consisting of 54 discrete features and 1 continuous feature. The anomaly proportion in MSL testing data is approximately 10.5\%. Soil Moisture Active Passive (SMAP) satellite contains 25 features, consisting of 24 discrete features and 1 continuous feature. The anomaly proportion in SMAP testing data is approximately 12.8\%.
\item Server Machine Dataset~\citep{su2019robust} is collected by large internet company. Server Machine Dataset (SMD) has 38 features, two of which are discrete and 36 of which are continuous. The anomaly proportion in SMD testing data is around 4.16\%.
\item Pooled Server Metrics~\citep{abdulaal2021practical} dataset is provided by eBay by capturing internally from application server nodes. Pooled Server Metrics (PSM) contains 25 features with no discrete feature. Among the testing data, the proportion of anomalies is approximately 27.76\%.
\end{itemize}

\begin{table}[t]
\caption{Summary of the datasets. Ratio (\%) represents the percentage of anomaly in the testing dataset.}
\begin{adjustbox}{width=\columnwidth}
\centering
\begin{tabular}{lcccccccc}
\hline
Dataset & \begin{tabular}[c]{@{}c@{}}Train\end{tabular} & Validation & Test & \begin{tabular}[c]{@{}c@{}}\# of Cont\end{tabular} ($c$) & \begin{tabular}[c]{@{}c@{}}\# of Disc\end{tabular} ($d$) & Ratio (\%)  \\
\hline
MSL & 46,655 &11,662  &73,729 & 1 &  54  &10.5\\
SMAP &108,148 &27,035  &427,617 & 1 &   24  &12.8 \\
SMD & 566,725 &141,680 &708,420 & 36 & 2  & 4.16\\
PSM & 103,289 &26,495  &87,841 & 25 & 0  & 27.76\\
\hline
\end{tabular}
\end{adjustbox}
\label{tbl:dataset}
\end{table}

\subsection{Baselines}
To compare the performance of the proposed model, we utilized several prediction-based time-series anomaly detection models and time-series forecasting models as baselines. 

\begin{itemize}
    \item LSTM-P~\citep{malhotra2015long} uses two-layer stacked LSTM network and a fully connected layer for final forecasting.
    \item DeepAnT~\citep{munir2018deepant} uses a CNN-based prediction model with two 1D convolution layers and two max pooling, and a fully connected layer.
    \item TCN-S2S-P~\citep{he2019temporal} applies a temporal convolutional network (TCN) with 1D dilated causal convolutions to time-series anomaly detection.
    \item MTAD-GAT~\citep{zhao2020multivariate} learns complex dependencies in time-series using two graph attention layers: temporal and feature dimensions.
    \item GDN~\citep{deng2021graph} is a graph-based model that has explainability for anomalies with structure learning and attention weights.
    \item GTA~\citep{chen2021learning} is a transformer-based model that learns a graph structure automatically and takes into consideration the long-term temporal dependencies.
    \item NLinear and DLinear~\citep{zeng2023transformers} are linear-based models that utilize temporal information with a linear layer. NLinear utilizes the normalization of time-series data, and DLinear utilizes the decomposition of time-series data.
    \item TimesNet~\citep{wu2022timesnet} is a Timesblock architecture based model, which incorporates a 2D backbone. It transforms 1D time series into a 2D space and analyzes the resulting 2D tensor using various 2D vision backbones. This allows it to effectively capture intra and interperiodic variations within the time series.
    \item PatchTST~\citep{nie2022time} is a transformer-based model that uses subseries-level patches of time series as input and has channel independence by processing multivariate time series as a single time series.
\end{itemize}

\subsection{Threshold Models}
We provide detailed information about GMM, ECOD, and DeepSVDD used as threshold models.
\begin{itemize}
     \item GMM is a density-based model that learns the density of the data and identifies points that do not fit well within that density as anomalies. The farther a data point deviates from the distribution, the lower the score it receives through GMM. Consequently, the lowest score value among the training data is used as the threshold.
     \item  ECOD~\citep{li2022ecod} is a density-based model, which learns the density of the data and considers points located in the both tail parts of the density distribution as anomalies. As a data point gets closer to the tail of the distribution, its score through ECOD tends to be higher. Hence, the threshold in ECOD-based anomaly detection is set to the highest score value among the training data.
     \item  DeepSVDD~\citep{ruff2018deep} is a boundary-based model to find the smallest hypersphere that includes the normal data on the latent feature space. If a data point is far outside the learned sphere, its score get higher and then that point will likely be classified to be an anomaly.
 \end{itemize}

\subsection{Evaluation Metrics}
To evaluate the time-series anomaly detection performance of our proposed model and baselines, we consider three evaluation metrics as follows: 
\begin{itemize}
    \item F1-@K~\citep{kim2022towards} first computes F1-score with the evaluation scheme in which all observations are considered correctly detected if the proportion of correctly detected anomalies in the consecutive anomaly segment exceeds the predefined criterion $K$. We use the metric F1-@K as the area under the curve of F1-score where $K$ varies by $0.1$ from $0$ to $1$ to mitigate the overestimation of point-adjusted F1-score~\citep{xu2018unsupervised} and the underestimation of point-wise F1-score. 
    \item F1-Composite~\citep{astha22evaluation} is calculated as the harmonic mean of point-wise precision and segment-wise recall for robust evaluation of segment-wise anomaly detection. 
    \item F1-Range~\citep{Wagner23timesead} considers time-series precision and recall with a set of actual anomaly segments and a set of predicted anomaly segments in order to overcome the problem that point-wise F1-score fails to discriminate predictive patterns by ignoring temporal dependencies. 
\end{itemize}

\begin{table*}[t]
     \caption{Components of our forecasting model.}
     \label{table:model}
\centering
\begin{subtable}[t]{0.48\textwidth}
        \centering
\begin{minipage}{\textwidth}
\begin{adjustbox}{width=\textwidth}
\begin{tabular}{cccc}
        \hline
        Layer & Design & Input size & Output size  \\ \hline
1 & $\text{Linear}_{trend}$ & 5 $\times$ $c$ & 1 $\times$ $c$  \\
2 & $\text{Linear}_{seasonality}$ & 5 $\times$ $c$  &  1 $\times$ $c$ \\ \hline
       \end{tabular}
\end{adjustbox}
\end{minipage}
\caption{Continuous model.}
\label{tab:model_cont}
\end{subtable}
\begin{subtable}[t]{0.48\textwidth}
        \centering
\begin{minipage}{\textwidth}
\begin{adjustbox}{width=\textwidth}
\begin{tabular}{cccc}
        \hline
Layer & Design & Input size & Output size  \\ \hline
1 & $\text{Linear}_{discrete}$ & 5 $\times$ $(d \times e)$ & 1 $\times$ $(d \times e)$ \\ \hline
        \end{tabular}
\end{adjustbox}
\end{minipage}
\caption{Discrete model.}
\label{tab:model_disc}
\end{subtable}

\vspace{1em}

\begin{subtable}{\textwidth}
    \centering
    \begin{tabular}{cccc}
    \hline
    Layer & Design & Input size & Output size  \\ \hline
    1 & $\text{Linear}_{input}$ & $(c+d)$ $\times$ $e$ & $h$ $\times$ $(c+d)$ $\times$ $e$ \\ 
    2 & $\text{Linear}_{output}$ &  $h$ $\times$ $(c+d)$ $\times$ $e$ &  $1$ $\times$ $(c+d)$ $\times$ $e$ \\
    3 & squeeze &  $1$ $\times$ $(c+d)$ $\times$ $e$ &  $(c+d)$ $\times$ $e$ \\\hline
    \end{tabular}
    \caption{Adaptive Graph Convolution Model.}
    \label{tab:model_agc}
\end{subtable}
\end{table*}

\section{Detailed Settings of Experiments}
We introduce the detailed setting of our experiments including model structure and the best hyperparameter. Here, we set the same sliding window size of 5 and prediction horizon length of 1 including baseline models. Additionally, we use the Adam~\citep{kingma2014adam} optimizer and set the learning rate to 0.005. Our forecasting models include continuous and discrete models, respectively, and an adaptive graph convolution model that extracts dependencies between features. Where $c$ and $d$ are the number of continuous and discrete features, $e$ is the one-hot embedding size of discrete features and $h$ is the hidden vector of the graph structure (cf. Table~\ref{table:model}). Finally, we use $\lambda$ for balancing between continuous features and discrete features in training loss. For each of the reported results, we list the best hyperparameter as follows:
\begin{itemize}
    \item For MSL, $c$ = 1, $d$ = 54, $h$ = 256, $e$ = 2, $\lambda$ = 1; 
    \item For SMAP, $c$ = 1, $d$ = 24, $h$ = 256, $e$ = 2, $\lambda$ = 1; 
    \item For SMD, $c$ = 36, $d$ = 2, $h$ = 256, $e$ = 16, $\lambda$ = 1; 
    \item For PSM, $c$ = 25, $e$ = 3, $h$ = 256; 
\end{itemize}

\section{Experiment Results}
In this section, we present the detailed results of our experiments in Table~\ref{tbl:results_msl} to~\ref{tbl:results_psm}.

\begin{wraptable}{r}{0.6\columnwidth}
\caption{Results of forecasting performance on the four benchmark datasets. Each value represents the mean of the evaluation metric and its standard deviation (Std).}
\label{tbl:results_mse3}
\centering
\begin{adjustbox}{width=\linewidth}
\begin{tabular}{c|c|c|c|c}
\hline
Dataset                                                                                               & \multicolumn{1}{c|}{MSL}  & \multicolumn{1}{c|}{SMAP} & \multicolumn{1}{c|}{SMD}  & PSM  \\ \hline
\begin{tabular}[c]{@{}c@{}}Mean Squared\\ Error (MSE)\end{tabular}                                                                           & \multicolumn{1}{c|}{Mean ± Std.} & \multicolumn{1}{c|}{Mean ± Std.} & \multicolumn{1}{c|}{Mean ± Std.} & Mean ± Std. \\ \hline
LSTM-P    &    1.7052 ± 0.0348                      &          0.0113 ± 0.0000                 &     0.0016 ± 0.0000                      &        0.0020 ± 0.0001                   \\
 DeepAnT   &    1.2387 ± 0.9148                      &     0.0314 ± 0.0105                      &      0.0040 ± 0.0002                     &          0.0014 ± 0.0005                  \\
 TCN-S2S-P &   1.0653 ± 0.1072                      &            0.0926 ± 0.0092                &        0.0233 ± 0.0026                    &           0.0027 ± 0.0000                \\
MTAD-GAT  &     1.8407 ± 0.0081                       &        \textbf{0.0105 ± 0.0005}                   &   0.0013 ± 0.0000            &             0.0016 ± 0.0002               \\
 GDN       &    20.824 ± 27.073                     &     0.0230 ± 0.0014                       &    0.0061 ± 0.0024                       &   0.0015 ± 0.0003                             \\
 GTA       &    1.8140 ± 0.0618                    &           0.0370 ± 0.0062                &        0.0186 ± 0.0007                   &         0.0042 ± 0.0004                    \\ \hline
NLinear   &   0.0671 ± 0.0001                      &       0.0188 ± 0.0000                    &            0.0013 ± 0.0000                &         \textbf{0.0001 ± 0.0000}                        \\
DLinear   &   0.8747 ± 0.0934                      &      0.0938 ± 0.0136                      &      0.0013 ± 0.0000                      &     \textbf{ 0.0001 ± 0.0000}                          \\
TimesNet  & \textbf{0.0278 ± 0.0513} &\textbf{0.0044 ± 0.0004}    &         0.0014  ±  0.0001   &     0.0010 ±	0.0003 
              \\ 
PatchTST  &    0.0650 ± 0.0028                       &       0.0124 ± 0.0002                    &       \textbf{0.0011 ± 0.0000}                    &      \textbf{ 0.0001 ± 0.0000}                         \\ \hline
\multicolumn{1}{c|}{Ours}                   &    \underline{0.0289 ± 0.0016}               &       \underline{0.0094 ± 0.0001}                    &        \underline{0.0013 ± 0.0000}                     &       \textbf{ 0.0001 ± 0.0000}                          \\ \hline
\end{tabular}
\end{adjustbox}
\vspace{-2em}
\end{wraptable}

\subsection{Forecasting Performance}
Table~\ref{tbl:results_mse3} shows the prediction performance of forecasting models. 
Since existing models consider discrete features as continuous features, our model also treats discrete features as continuous features during the evaluation process and only uses mean square error (MSE).

Evaluating the model only with MSE is disadvantageous to our model, which trained discrete features with cross-entropy (CE) loss. Nonetheless, our forecasting model shows comparable or better performance than the latest time-series forecasting models. Additionally, we will show that evaluating predictive performance using MSE does not guarantee the performance of proactive anomaly detection in the next section. 

\subsection{Anomaly Detection Performance}
Diff represents the difference between the Ground Truth and the result of using the predicted value as an input to the threshold model. The closer to 0, the more similar the prediction to the ground truth value. For models that did not predict well and judged all samples to be anomalies, the Diff value is displayed as -. In some cases, there were models with a lower Diff than our forecasting model, but those models had a much greater variance in forecasting performance than our forecasting model (cf. Table~\ref{tbl:results_mse3}). Therefore, we can find the anomaly in advance through accurate prediction.

For a detailed comparison, we reported a comprehensive analysis by presenting the mean and standard deviation derived from five repeated experiments in Table~\ref{tbl:Relability result}. The ground truth in the last row of Table~\ref{tbl:Relability result} is the evaluation of testing data by the data-driven model trained with training data. In other words, if the predicted values of each model are similar to the testing data, they are identical to the ground truth. Therefore, the goal of our experiment is for the evaluation scores of predicted values, as determined by the data-driven model, to closely resemble those of the ground truth values.

\begin{table*}[t]
\caption{Mean of evaluation metric and its standard deviation (std). Each value is represented as mean ± std. Ground Truth is the result of fitting the testing data to the trained data-driven model. Bold is the most similar performance to Ground Truth.}
\label{tbl:Relability result}
\setlength{\tabcolsep}{2pt}
\begin{adjustbox}{width=\textwidth}
\begin{tabular}{cc|ccc|ccc|ccc}
\hline
\multicolumn{2}{c|}{Dataset-Data-driven Model}                                         & \multicolumn{3}{c|}{MSL-GMM}                                         & \multicolumn{3}{c|}{SMAP-ECOD}                                           & \multicolumn{3}{c}{PSM-DeepSVDD}                                               \\ \hline
\multicolumn{2}{c|}{Metric}                                                                                                & F1-@K                 & F1-C                  & F1-R                  & F1-@K                 & F1-C                     & F1-R                  & F1-@K                    & F1-C                     & F1-R                     \\ \hline
\multicolumn{1}{c|}{\multirow{6}{*}{\begin{tabular}[c]{@{}c@{}}Unsupervised\\ Time-series  \\ Anomaly Model\end{tabular}}} & LSTM-P    & 0.1906 ± 0.0000         & 0.1906   ± 0.0000       & 0.1905 ± 0.0000        & 0.2031 ± 0.0146       & 0.1792 ± 0.0124          & 0.1491 ± 0.0171       &  0.4571 ± 0.0031 & \textbf{0.4347 ± 0.0024}  & 0.4304 ± 0.0024 \\
\multicolumn{1}{c|}{}         & DeepAnT   & 0.0451 ± 0.0026      & \textbf{0.2179 ± 0.0323}       &  0.0404 ± 0.0284      & 0.1981 ± 0.0276       & 0.2040 ± 0.0217 & 0.1669 ± 0.0163       & 0.4572 ± 0.0017 & 0.4358 ± 0.0012 & 0.4316 ± 0.0014\\
\multicolumn{1}{c|}{} & TCN-S2S-P & 0.1906 ± 0.0000          & 0.1906 ± 0.0000          & 0.1905 ± 0.0000         & 0.1284 ± 0.0143       & 0.1464 ± 0.0060           & 0.0918 ± 0.0128       & 0.4505 ± 0.0033 & 0.4307 ± 0.0043 & 0.4241 ± 0.0049 \\
\multicolumn{1}{c|}{}    & MTAD-GAT  & 0.1906 ± 0.0000       & 0.1906 ± 0.0000       & 0.1905 ± 0.0000       & 0.1896 ± 0.0070        & 0.1749 ± 0.0050          & 0.1547 ± 0.0045       &  0.4560 ± 0.0021 & 0.4308 ± 0.0041 & 0.4263 ± 0.0037 \\
\multicolumn{1}{c|}{}  & GDN       & 0.1906 ± 0.0000          & 0.1906 ± 0.0000          & 0.1905 ± 0.0000          & 0.1351 ± 0.0082       & 0.1413 ± 0.0012          & 0.0968 ± 0.0058       & 0.4601 ± 0.0090 & 0.4341 ± 0.0010 & 0.4262 ± 0.0033 \\
\multicolumn{1}{c|}{}  & GTA       & 0.1906 ± 0.0000       & 0.1906 ± 0.0000       & 0.1905 ± 0.0000       & 0.2348 ± 0.0230       & 0.2020 ± 0.0164          & 0.1731 ± 0.0159       &   0.4631 ± 0.0059 & 0.4428 ± 0.0059 & 0.4357 ± 0.0065      \\ \hline
\multicolumn{1}{c|}{\multirow{4}{*}{\begin{tabular}[c]{@{}c@{}}Time-series\\ Prediction Model\end{tabular}}}   & NLinear   & 0.2451 ± 0.0000          & 0.1996 ± 0.0000          & 0.1858 ± 0.0000         & 0.0435 ± 0.0000          & 0.1989 ± 0.0001          & 0.0208 ± 0.0000          & \textbf{0.4458 ± 0.0001} & 0.4332 ± 0.0001 & 0.4311 ± 0.0002      \\
\multicolumn{1}{c|}{}   & DLinear   & 0.1906 ± 0.0000       & 0.1906 ± 0.0000       & 0.1905 ± 0.0000       & 0.0426 ± 0.0004       & 0.1990 ± 0.0055          & 0.0194 ± 0.0008       &   0.4459 ± 0.0000 & 0.4331 ± 0.0001 & 0.4312 ± 0.0002       \\
\multicolumn{1}{c|}{}       & TimesNet  &    0.1906 ± 0.0000    & 0.1906 ± 0.0000 & 0.1905 ± 0.0000    &   0.3048 ± 0.0525 & \textbf{0.2684 ± 0.0484} & 0.2153 ± 0.0389 &   0.4511 ± 0.0021 & 0.4369 ± 0.0031 & 0.4337 ± 0.0028 \\
\multicolumn{1}{c|}{}       & PatchTST & 0.1906 ± 0.0000          & 0.1906 ± 0.0000          & 0.1905 ± 0.0000          & 0.0423 ± 0.0006       & 0.2021 ± 0.0047          & 0.0194 ± 0.0009       &  0.4459 ± 0.0001 & 0.4346 ± 0.0001 & \textbf{0.4326 ± 0.0002}   \\ \hline

\multicolumn{2}{c|}{OURS}    & \textbf{0.0837 ± 0.0014} & 0.1972 ± 0.0001 & \textbf{0.0857 ± 0.0004} & \textbf{0.0210 ± 0.0000} & 0.1650 ± 0.0000             & \textbf{0.0014 ± 0.0000} & 0.4460 ± 0.0002 & 0.4350 ± 0.0001 & 0.4331 ± 0.0001  \\ \hline
\multicolumn{2}{c|}{Ground Truth}             & 0.0916               & 0.3428                & 0.0937                & 0.0290                & 0.2908                   & 0.0046                &  0.4437 & 0.4347 & 0.4322        \\ \hline
\end{tabular}
\end{adjustbox}
\end{table*}

When GMM is used as the data-driven model in the MSL dataset (MSL-GMM), our model shows the best anomaly detection performance. In addition, all the models except for DeepAnT, NLinear, and ours, predict all samples as anomalies, showing the same performance. In SMAP-ECOD, our model also shows the best and the most consistent performance. These results also imply that the prediction performance of our model has low variability, as shown in most of the results. In PSM-DeepSVDD, our model does not show the best performance, but the performance difference with the best model is very small, up to 0.0005.

\begin{wrapfigure}{r}{8cm}
\vspace{-1em}
\centering
  \begin{subfigure}{\columnwidth}
    {{\includegraphics[width=0.3\columnwidth]{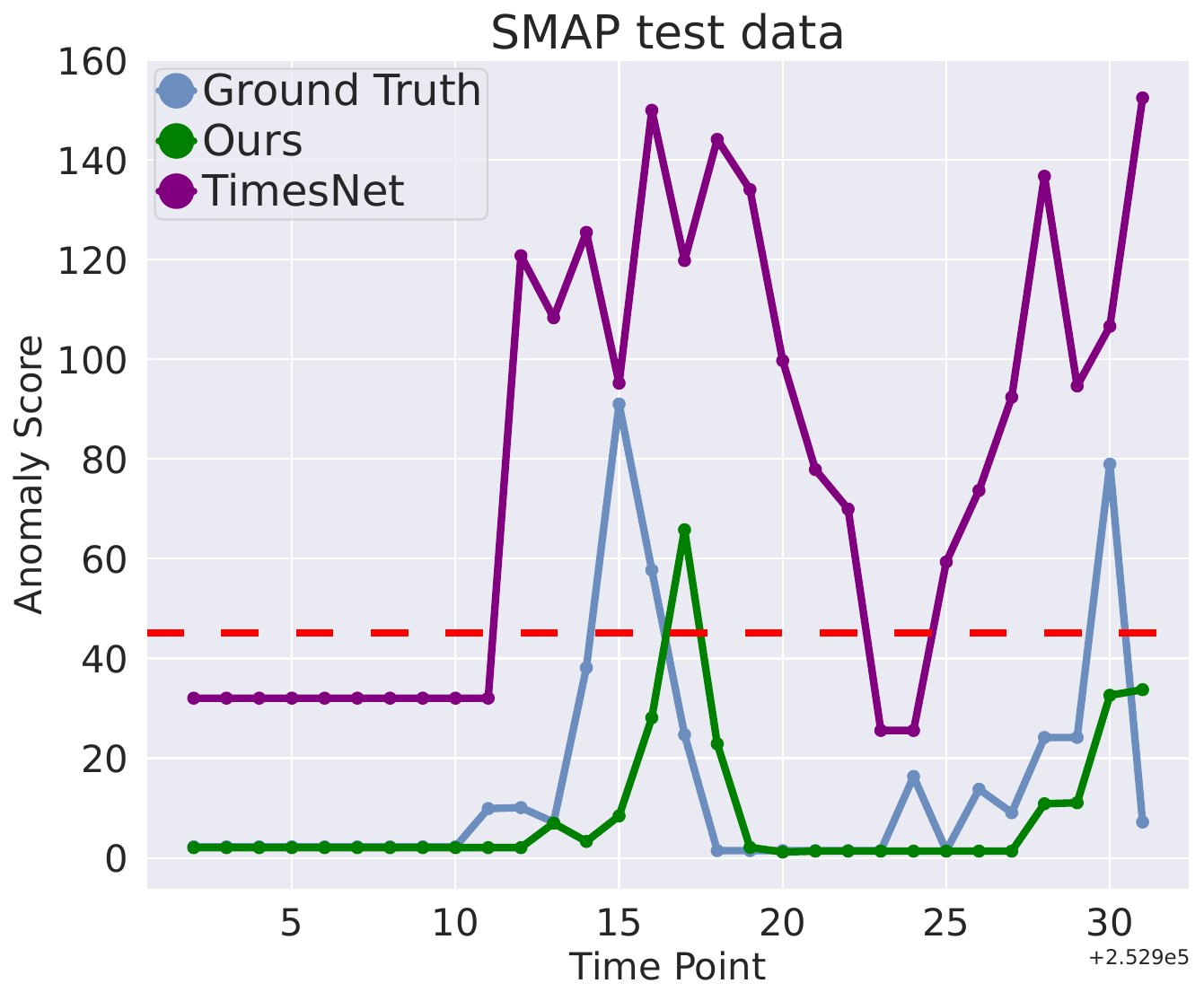}}{\includegraphics[width=0.3\columnwidth]{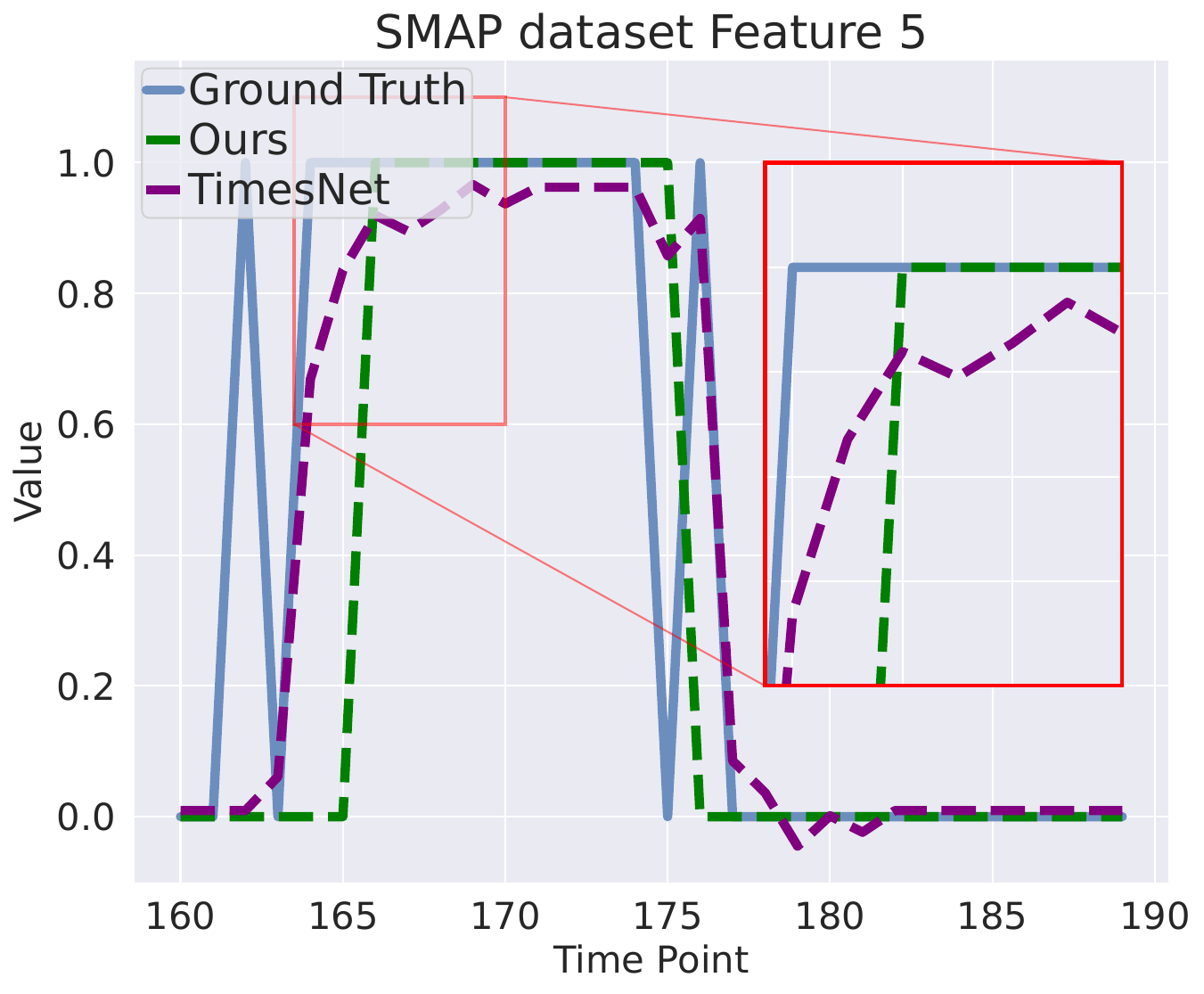}}}
  \end{subfigure}
\caption{Visualization of comparison between ours and TimeNet in the SMAP dataset. \textit{Left}: Anomaly score by the trained ECOD. \textit{Right}: Predicted values for the categorical feature.}
\vspace{-2em}
\label{fig:casestudy1}
\end{wrapfigure}

In Table~\ref{tbl:Relability result}, our model shows better performance in terms of F1-@K and F1-Range when evaluated on MSL-GMM and SMAP-ECOD. However, it shows worse performance in the F1-Composite. To analyze these differences, we visualize the evaluation of predicted values by a trained data-driven model. Figure~\ref{fig:casestudy1} (left)  reveals that there is a big difference between the ground truth and predicted values of TimesNet. As a result, TimesNet identifies a majority of the samples as anomalies including normal samples within testing data, showing better performance than our model F1-Composite in Table~\ref{tbl:Relability result}.

Although  TimesNet shows better prediction performance in terms of MSE compared to our model, our model shows more similar anomaly scores for each time point than TimesNet. As visualized in the right panel of Figure~\ref{fig:casestudy1}, TimesNet produces continuous values for discrete features, while our model predicts properly discrete values. As a result, even though our model shows worse forecasting performance in terms of MSE, our model forecasts values that are more closely aligned with the ground truth compared to other models.

\begin{table*}[ht]
\caption{Results of anomaly detection experiment on MSL dataset. When the data-driven model decides all samples to be abnormal, the F1 score is 0.1906.}
\label{tbl:results_msl}
\begin{adjustbox}{width=\textwidth}
\begin{tabular}{cc|cc|cc|cc|cc|cc|cc|cc|cc|cc}
\hline
\multicolumn{2}{c|}{\begin{tabular}[c]{@{}c@{}}Data-driven\\ Model\end{tabular}}                                           & \multicolumn{6}{c|}{GMM}    & \multicolumn{6}{c|}{ECOD}   & \multicolumn{6}{c}{DeepSVDD}     \\ \hline
\multicolumn{2}{c|}{Metrics}            & F1-@K  & Diff & F1-C  & Diff  & F1-R  & Diff & F1-@K  & Diff & F1-C  & Diff  & F1-R  & Diff & F1-@K  & Diff & F1-C  & Diff  & F1-R  & Diff \\ \hline
\multicolumn{1}{c|}{\multirow{6}{*}{\begin{tabular}[c]{@{}c@{}}Unsupervised\\ Time-series  \\ Anomaly Model\end{tabular}}} & LSTM-P     &   0.1906 &	-	&	 0.1906 &	-	&	 0.1905 &	-	&	 0.1906 &	-	&	 0.1906 &	-	&	 0.1905 &	-	&	 0.2026 &	0.0296	&	 0.1982 &	0.0511	&	 0.1973 &	0.0562	\\

\multicolumn{1}{c|}{}                                                                                        & DeepAnT    & 0.0451  &	0.0465	&	 \textbf{0.2179 } 	&	\textbf{0.1249	}&	 0.0404 &	0.0533	&	 0.1906 &	-	&	 0.1906 &	-	&	 0.1905 	&	-	&	 0.1906 &	0.0176	&	 0.1906 &	0.0435	&	 0.1905 	&	0.0494	\\

\multicolumn{1}{c|}{}                                                                                           & TCN-S2S-P  &   0.1906   &    -    & 0.1906 &  - & 0.1905 &  -  & 0.1906  &    -  & 0.1906  &  -  & 0.1905  &   -  & 0.1908 &   0.0178   & 0.1908 &    0.0437   & 0.1907 & 0.0496  \\
\multicolumn{1}{c|}{}                                                                                           & MTAD-GAT  &  0.1906 &	-	&	 0.1906 &	-	&	 0.1905 &	-	&	 0.1906 &	-	&	 0.1906 &	-	&	 0.1905 &	-	&	 0.2007  &	0.0277	&	\textbf{ 0.1846  }	&	\textbf{0.0375}	&	 \textbf{0.1812 } 	&\textbf{0.0401}	\\

\multicolumn{1}{c|}{}                                                                                           & GDN       &        0.1906 &	-	&	 0.1906 &	-	&	 0.1905 &	-	&	 0.1906 &	-	&	 0.1906 &	-	&	 0.1905 &	-	&	 0.1970 &	0.0240	&	 0.1958 &	0.0487	&	 0.1953 &	0.0542	\\

\multicolumn{1}{c|}{}                                                                                           & GTA        &      0.1906 &	-&	 0.1906 &	-	&	 0.1905 &	-	&	 0.1906 &	-	&	 0.1906 &	-	&	 0.1905 &	-	&	 0.1913 &	0.0183	&	 0.1910 &	0.0439	&	 0.1908 &	0.0497	\\  \hline

\multicolumn{1}{c|}{\multirow{4}{*}{\begin{tabular}[c]{@{}c@{}}Time-series\\  Prediction Model\end{tabular}}}   & NLinear    &   0.2451 &	0.1535	&	 0.1996 &	0.1432	&	 0.1858 &	0.0921	&	 0.1906 &	-	&	 0.1906 &	-	&	 0.1905 &	-	&	 0.2222 &	0.0492	&	 0.2178 &	0.0707	&	 0.2150 &	0.0739	\\

\multicolumn{1}{c|}{}   & DLinear    &      0.1906 &	-	&	 0.1906 &	-	&	 0.1905 &	-	&	 0.1906 &	-	&	 0.1906 &	-	&	 0.1905 &	-	&	 0.2213 &	0.0483	&	 0.2169 &	0.0698	&	 0.2142 &	0.0731	\\

\multicolumn{1}{c|}{}   & TimesNet &   0.1906 &	-	&	 0.1906 &	- &	 0.1905 &	- &	 0.1906  &	-	&	 0.1906  &	-	&	 0.1905 &	- & 0.2048 & 0.0318& 0.2013& 0.0542& 0.2002& 0.0591\\

\multicolumn{1}{c|}{}  & PatchTST    &   0.1906 &	-	&	 0.1906 &	- &	 0.1905 &	- &	 0.1906  &	-	&	 0.1906  &	-	&	 0.1905 &	-	&	 0.2014 &	0.0284	&	 0.1996 &	0.0525	&	 0.1987 &	0.0576	\\ \hline
\multicolumn{2}{c|}{Ours}                                                                                                    &    \textbf{0.0837 } 	&	\textbf{0.0079}	&	 0.1972  &	0.1456	&	\textbf{ 0.0853 } 	&	\textbf{0.0084}&	 \textbf{0.2279 }	&\textbf{	0.2136}	&	\textbf{ 0.1940  }	&	\textbf{0.0440}	&	\textbf{ 0.1774 }	&	\textbf{0.1750}	&	 \textbf{0.1902 }	&	\textbf{0.0172}	&	 0.1870 &	0.0399	&	 0.1869 &	0.0458	\\
\hline

\multicolumn{2}{c|}{Ground Truth} &0.0916	&&	0.3428	&&	0.0937	&&	0.0143	&&	0.1500	&&	0.0024	&&	0.1730	&&	0.1471	&&	0.1411
\\\hline
\end{tabular}
\end{adjustbox}
\end{table*}

\begin{table*}[ht]
\caption{Results of anomaly detection experiment on SMAP dataset. When the data-driven model decides all samples to be abnormal, the F1 score is 0.2268.}
\label{tbl:results_smap}
\begin{adjustbox}{width=\textwidth}
\begin{tabular}{cc|cc|cc|cc|cc|cc|cc|cc|cc|cc}
\hline
\multicolumn{2}{c|}{\begin{tabular}[c]{@{}c@{}}Data-driven\\ Model\end{tabular}}                                           & \multicolumn{6}{c|}{GMM}    & \multicolumn{6}{c|}{ECOD}   & \multicolumn{6}{c}{DeepSVDD}     \\ \hline
\multicolumn{2}{c|}{Metrics}            & F1-@K  & Diff & F1-C  & Diff  & F1-R  & Diff & F1-@K  & Diff & F1-C  & Diff  & F1-R  & Diff & F1-@K  & Diff & F1-C  & Diff  & F1-R  & Diff \\ \hline
\multicolumn{1}{c|}{\multirow{6}{*}{\begin{tabular}[c]{@{}c@{}}Unsupervised\\ Time-series  \\ Anomaly Model\end{tabular}}} & LSTM-P     &   0.2269 &	0.1902	&	 0.2268 &	-	&	 0.2268 &	-	&	 0.2031 &	0.1741	&	 0.1792 &	0.1116	&	 0.1491 	&	0.1445	&	 0.2501 &	0.0020	&	 0.2393 &	0.0034	&	 0.2332 &	0.0032	\\

\multicolumn{1}{c|}{}                                                                                           & DeepAnT    &   0.0001 &	0.0366	&	 0.0059 &	0.3126	&	 0.0 &	0.0078	&	 0.1981 &	0.1691	&	0.2040	&	0.0868	&	 0.1669 &	0.1623	&	 0.2268 &	0.0253	&	 0.2268 &	0.0159	&	 0.2268 &	0.0096	\\

\multicolumn{1}{c|}{}                                                                                           & TCN-S2S-P  &   0.2268 & - & 0.2268 &  - & 0.2268 &   - & 0.1284 &  0.0994   & 0.1464 & 0.1444  & 0.0918 & 0.0872 &\textbf{0.2528  }  &   \textbf{0.0007}  & \textbf{0.2438 } & \textbf{0.0011}  &\textbf{ 0.2374  } & \textbf{0.0010}   \\

\multicolumn{1}{c|}{}                                                                                           & MTAD-GAT  &   0.2281 &	0.1914	&	 0.2266	& 0.0919	&	 0.2263 &	0.2185	&	 0.1896 &	0.1606	&	 0.1749 &	0.1159	&	 0.1547 	&	0.1501	&	 0.2491 &	0.0030	&	0.2394 &	0.0033	&	 0.2334  &	0.0030	\\
\multicolumn{1}{c|}{}          & GDN     &  0.2268  &	-	&	 0.2268 &	-	&	 0.2268 &	-	&	 0.1351 &	0.1061	&	 0.1413 &	0.1495	&	 0.0968 &	0.0922	&	 0.2385 &	0.0136	&	 0.2319 &	0.0108	&	 0.2284 &	0.0080	\\
\multicolumn{1}{c|}{}    &  GTA &    0.2277 &	0.1910	&	 0.2268 &	-	&	 0.2269 &	0.2191	&	 0.2348 &	0.2058	&	 0.2020 &	0.0888	&	 0.1731 &	0.1685	&	 0.2717 &	0.0196	&	 0.2540 &	0.0113	&	 0.2444 &	0.0080	\\ \hline
\multicolumn{1}{c|}{\multirow{4}{*}{\begin{tabular}[c]{@{}c@{}}Time-series\\  Prediction Model\end{tabular}}}   & NLinear    &  0.1373 &	0.1006	&	 0.1402 &	0.1783	&	 0.0979  &	0.0901	&	 0.0435 &	0.0145	&	 0.1989 &	0.0919	&	 0.0208 &	0.0162	&	 0.2315 &	0.0206	&	 0.2258 &	0.0169	&	 0.2229 &	0.0135	\\
\multicolumn{1}{c|}{}   & DLinear    &     0.2268  &	-	&	 0.2268 &	-	&	 0.2268 &	-	&	 0.0426 &	0.0136	&	 0.1990  &	0.0918	&	 0.0194 &	0.2222	&	 0.2309  &	0.0212	&	 0.2254 &	0.0173	&	 0.2225 &	0.0139	\\

\multicolumn{1}{c|}{}   & TimesNet    & 0.2358     &	0.1991 &	\textbf{ 0.2305}  &	\textbf{ 0.0880}	&	0.2273 &	0.2195	& 0.3048  &	0.2758 &	\textbf{0.2684} &	\textbf{0.0224}	&	0.2153 &	0.2107	&	0.2495  & 0.0026 &	0.2386  &	0.0041 & 0.2322	 & 0.0042 \\
\multicolumn{1}{c|}{}                                                                                           & PatchTST    &   0.2268 &	-	&	 0.2268 &	-	&	 0.2268 &	-	&	 0.0423 &	0.0133	&	 0.2021 &	0.0887	&	 0.0194 &	0.0148	&	 0.2282  	&	0.0239	&	 0.2234 &	0.0193	&	 0.2209 &	0.0155	\\\hline
\multicolumn{2}{c|}{Ours}                                                                                                    &  \textbf{0.0185 }	&	\textbf{0.0182}	&	 0.2181  &	0.1004	&	\textbf{ 0.0026 } 	&	\textbf{0.0052}	&	 \textbf{0.0210 }	&	\textbf{0.0080}	&	 0.1650  &	0.1258	&	 \textbf{0.0014 } 	&	\textbf{0.0032}	&	 0.2416  &	0.0105	&	 0.2353 &	0.0074	&	 0.2314 &	0.0050	\\
 \hline
\multicolumn{2}{c|}{Ground Truth} & 0.0367	&&	0.3185	&&	0.0078	&&	0.0290	&&	0.2908	&&	0.0046	&&	0.2521	&&	0.2427	&&	0.2364 \\\hline
\end{tabular}
\end{adjustbox}
\end{table*}

\begin{table*}[ht]
\caption{Results of anomaly detection experiment on SMD dataset. When the data-driven model decides all samples to be abnormal, the F1 score is 0.0806.}
\label{tbl:results_smd}
\begin{adjustbox}{width=\textwidth}
\begin{tabular}{cc|cc|cc|cc|cc|cc|cc|cc|cc|cc}
\hline
\multicolumn{2}{c|}{\begin{tabular}[c]{@{}c@{}}Data-driven\\ Model\end{tabular}}                                           & \multicolumn{6}{c|}{GMM}    & \multicolumn{6}{c|}{ECOD}   & \multicolumn{6}{c}{DeepSVDD}     \\ \hline
\multicolumn{2}{c|}{Metrics}            & F1-@K  & Diff & F1-C  & Diff  & F1-R  & Diff & F1-@K  & Diff & F1-C  & Diff  & F1-R  & Diff & F1-@K  & Diff & F1-C  & Diff  & F1-R  & Diff \\ \hline
\multicolumn{1}{c|}{\multirow{6}{*}{\begin{tabular}[c]{@{}c@{}}Unsupervised\\ Time-series  \\ Anomaly Model\end{tabular}}} & LSTM-P     &     \textbf{ 0.1240 }	&	\textbf{0.0204}	&	 0.1077 &	0.0833	&	 0.1050 &	0.0285	&	 0.1468  &	0.1077	&	 0.2125 	&	0.1037	&	 0.0636  &	0.0526	&	 0.0920 &	0.0024	&	 0.0763 &	0.0018	&	 0.0757 &	0.0016	\\
\multicolumn{1}{c|}{}                                                                                           & DeepAnT    &   0.1131 	&	0.0313	&	 0.0979 &	0.0931	&	 0.0554  &	0.0211	&	 0.0424  &	0.0033	&	 0.0637 &	0.0451	&	 0.0124 &	0.0014	&	\textbf{ 0.0928} 	&	\textbf{0.0016}	&	 0.0754 &	0.0027	&	 0.0746 &	0.0027	\\

\multicolumn{1}{c|}{}                                                                                           & TCN-S2S-P  &0.0832  &  0.0612   & 0.0811 &  0.1099 & 0.0809  &  0.0044 &  0.0203  &  0.0188 &  0.0612   &  0.0476 &  0.0033  &   0.0077  & 0.0924 &   0.0020    & 0.0766 &  0.0015 &  0756 &   0.0017  \\
\multicolumn{1}{c|}{}                                                                                           & MTAD-GAT  &   0.2176  &	0.0732	&	 0.1948	&	0.0038	&	 0.1662 &	0.0897	&	 0.1407 &	0.1016	&	 0.2046 &	0.0958	&	 0.0497 &	0.0387	&	 0.0908 &	0.0036	&	 0.0743 &	0.0038	&	 0.0736 &	0.0037	\\
\multicolumn{1}{c|}{}                                                                                           & GDN       & 0.1180 &	0.0264	&	 0.1051 &	0.0859	&	 0.1031 &	0.0266	&	 0.1650 &	0.1259	&	 0.2091 &	0.1003	&	 0.0726 &	0.0616	&	 0.0925 &	0.0019	&	\textbf{ 0.0777 }	&	\textbf{0.0004}	&	\textbf{ 0.0770 }	&	\textbf{0.0003}	\\
\multicolumn{1}{c|}{}                                                                                           & GTA & 0.1047  &	0.0397	&	 0.0845 &	0.1065	&	\textbf{ 0.0768 } 	&	\textbf{0.0003}	&	 0.0006 &	0.0385	&	 0.0011 &	0.1077	&	 0.0001 &	0.0109	&	 0.0901 &	0.0043	&	 0.0757 &	0.0024	&	 0.0749 &	0.0024	\\
\hline
\multicolumn{1}{c|}{\multirow{4}{*}{\begin{tabular}[c]{@{}c@{}}Time-series\\  Prediction Model\end{tabular}}}   & NLinear    &      0.1148  &	0.0295	&	 0.1208 &	0.0702	&	 0.0540 &	0.0225	&	 0.0427  &	0.0036	&	 0.0807  &	0.0281	&	 0.0126 &	0.0016	&	 0.0918 &	0.0026	&	 \textbf{0.0785 } 	&	\textbf{0.0004}	&	 0.0778 &	0.0005	\\
\multicolumn{1}{c|}{}                                                                                           & DLinear  & 0.0917  &	0.0527	&	 0.1062 &	0.0848	&	 0.0494 &	0.0271	&	 0.0358 &	0.0033	&	 0.0660 &	0.0428	&	 \textbf{0.0103} 	&	\textbf{0.0007}	&	 0.0917 &	0.0027	&	\textbf{ 0.0785 }	&	\textbf{0.0004}	&	 0.0778 &	0.0005	\\
\multicolumn{1}{c|}{}                                                                                           & TimesNet  & 	0.1972 & 0.0528 & \textbf{0.1921}& \textbf{0.0011} & 0.1483 & 0.0718 &  0.1225 & 0.0834 & 0.2242 & 0.1154 & 0.0581 & 0.0471 & 0.0915 & 0.0029 & 0.0757 & 0.0024 & 0.0751 & 0.0022 \\
\multicolumn{1}{c|}{}                                                                                           & PatchTST    &   0.1190 &	0.0254	&	 0.1489 &	0.0421	&	 0.0699 &	0.0066	&	\textbf{ 0.0410 }	&	\textbf{0.0019}	&	 0.0757 &	0.0331	&	 0.0121 &	0.0011	&	 0.0910  &	0.0034	&	 0.0773 &	0.0008	&	 0.0767 &	0.0006	\\
 \hline
\multicolumn{2}{c|}{Ours}                                                                                                    &   0.1996 &	0.0552	&	 0.2178 &	0.0268	&	 0.1278 &	0.0513	&	 0.0463 &	0.0072	&	\textbf{ 0.0865 }	&	\textbf{0.0223}	&	 0.0131 &	0.0021	&	 0.0924 &	0.0020	&	 0.0787 &	0.0006	&	 0.0780 &	0.0007	\\ \hline
\multicolumn{2}{c|}{Ground Truth} & 0.1444	&&	0.1910	&&	0.0765	&&	0.0391	&&	0.1088	&&	0.0110	&&	0.0944	&&	0.0781	&&	0.0773\\\hline
\end{tabular}
\end{adjustbox}
\end{table*}

\begin{table*}[ht]
\caption{Results of anomaly detection experiment on PSM dataset. When the data-driven model decides all samples to be abnormal, the F1 score is 0.4351.}
\label{tbl:results_psm}
\begin{adjustbox}{width=\textwidth}
\begin{tabular}{cc|cc|cc|cc|cc|cc|cc|cc|cc|cc}
\hline
\multicolumn{2}{c|}{\begin{tabular}[c]{@{}c@{}}Data-driven\\ Model\end{tabular}}                                           & \multicolumn{6}{c|}{GMM}    & \multicolumn{6}{c|}{ECOD}   & \multicolumn{6}{c}{DeepSVDD}     \\ \hline
\multicolumn{2}{c|}{Metrics}            & F1-@K  & Diff & F1-C  & Diff  & F1-R  & Diff & F1-@K  & Diff & F1-C  & Diff  & F1-R  & Diff & F1-@K  & Diff & F1-C  & Diff  & F1-R  & Diff \\ \hline
\multicolumn{1}{c|}{\multirow{6}{*}{\begin{tabular}[c]{@{}c@{}}Unsupervised\\ Time-series  \\ Anomaly Model\end{tabular}}} & LSTM-P     &     0.0034 &	0.0452	&	 0.0160 &	0.1128	&	 0.0007 	&	0.0139	&	 0.0016  &	0.0228	&	 0.0219 &	0.0581	&	 0.0013 &	0.0059	&	 0.4571 &	0.0134	&	 \textbf{0.4347  }	&	\textbf{0}	&	 0.4304 	&	0.0018	\\

\multicolumn{1}{c|}{}                                                                                           & DeepAnT    &    0.0 	&	0.0486	&	 0.0 &	0.1288	&	 0.0 &	0.0146	&	 0.0087 	&	0.0157	&	 0.0590 &	0.0210	&	 0.0057 &	0.0015	&	 0.4572 	&	0.0135	&	 0.4358  &	0.0011	&	 0.4316 &	0.0006	\\
\multicolumn{1}{c|}{}                                                                                           & TCN-S2S-P  &    0.0  &  0.0486  & 0.0 &  0.1288   &0.0  &   0.0146  & 0.0 & 0.0244 & 0.0  &  0.0800  & 0.0 & 0.0072  &0.4505 &  0.0068  & 0.4307  &  0.0040   & 0.4241 &  0.0081  \\
\multicolumn{1}{c|}{}                                                                                           & MTAD-GAT  &   0.0  &	0.0486	&	 0.0055  &	0.1233	&	 0.0006 &	0.0140	&	 0.0090 &	0.0154	&	 0.0434 &	0.0366	&	 0.0025 &	0.0047	&	 0.4560 &	0.0123	&	 0.4308  &	0.0039	&	 0.4263 &	0.0059	\\
\multicolumn{1}{c|}{}                                                                                           & GDN       &  \textbf{ 0.0322 } 	&	\textbf{0.0164}	&	 \textbf{0.0691 } 	&	\textbf{0.0597}	&	\textbf{ 0.0163 }	&\textbf{	0.0017}	&	 0.0145  &	0.0099	&	 0.0799 &	0.0001	&	 0.0080  &	0.0008	&	 0.4601 &	0.0164	&	 0.4341 &	0.0006	&	 0.4262 &	0.0060	\\

\multicolumn{1}{c|}{}                                                                                           & GTA        &   0.0 &	0.0486	&	 0.0 &	0.1288	&	 0.0  &	0.0146	&	 0.0005 	&	0.0239	&	 0.0055 &	0.0745	&	 0.0002 &	0.0070	&	 0.4631 &	0.0194	&	 0.4428 &	0.0081	&	 0.4357 &	0.0035	\\
\hline
\multicolumn{1}{c|}{\multirow{4}{*}{\begin{tabular}[c]{@{}c@{}}Time-series\\  Prediction Model\end{tabular}}}   & NLinear    &    0.0250 &	0.0236	&	 0.0539 &	0.0749	&	 0.0015 &	0.0131	&	 0.0241 &	0.0003	&	 \textbf{0.0800 } 	&	\textbf{0}	&	\textbf{ 0.0072 } 	&	\textbf{0}	&	 \textbf{0.4458 }	&	\textbf{0.0021}	&	 0.4332 &	0.0015	&	 0.4311 &	0.0011	\\
\multicolumn{1}{c|}{}                                                                                           & DLinear & 0.0181 &	0.0305	& 0.0540 &	0.0747 & 0.0010  &	0.0136	& 0.0239 &	0.0005	&	\textbf{0.0800 }&	\textbf{0}	& 0.0069 &	0.0003	&	 0.4459 &	0.0022	&	 0.4331 &	0.0016	&	 0.4312 &	0.0010	\\

\multicolumn{1}{c|}{}   & TimesNet    & 0.0537& 0.0051 & 0.0160 &	0.1128 & 0.0036 & 0.011 & 0.017 & 0.0074 &	0.0642 & 0.0158 & 0.0038 &	0.0034 & 0.4511 & 0.0074 & 0.4369 & 0.0022 & 0.4337 & 0.0015
\\
\multicolumn{1}{c|}{}                                                                                           & PatchTST    &    0.0244 &	0.0242	&	 0.0274 &	0.1014	&	 0.0006 &	0.0140	&	 0.0241  &	0.0003	& \textbf{0.0800 }	&	\textbf{0}	&	 0.0070 &	0.0002	&	 0.4459 &	0.0022	& 0.4346 &	0.0001	&	 \textbf{0.4326 } 	&	\textbf{0.0004}	\\
\hline
\multicolumn{2}{c|}{Ours}                                                                                                    &   0.0244 &	0.0242	&	 0.0591  &	0.0697	& 0.0020  &	0.0126	&	\textbf{ 0.0243 }	&	\textbf{0.0001}	&	\textbf{ 0.0800 } 	&	\textbf{0	}&	 0.0073 &	0.0001	&	 0.4460  &	0.0023	&	 0.4350 &	0.0003	& 0.4331 &	0.0009	\\
 \hline
\multicolumn{2}{c|}{Ground Truth} & 0.0486	&&	0.1288	&&	0.0146	&&	0.0244	&&	0.0800	&&	0.0072	&&	0.4437	&&	0.4347	&&	0.4322
\\\hline
\end{tabular}
\end{adjustbox}
\end{table*}

\subsection{Ablation Studies}
For more accurate prediction performance, our forecasting model consists of various factors. Among them, the most important factors are graph structure and separate training processes of continuous and discrete features. Therefore, we investigate the role and effectiveness of each component through ablation studies of prediction performance using the SMD dataset.

\subsubsection{Graph Structure.}
\begin{wrapfigure}{r}{8cm}
\vspace{-1em}
\centering
{\includegraphics[width=0.5\columnwidth]{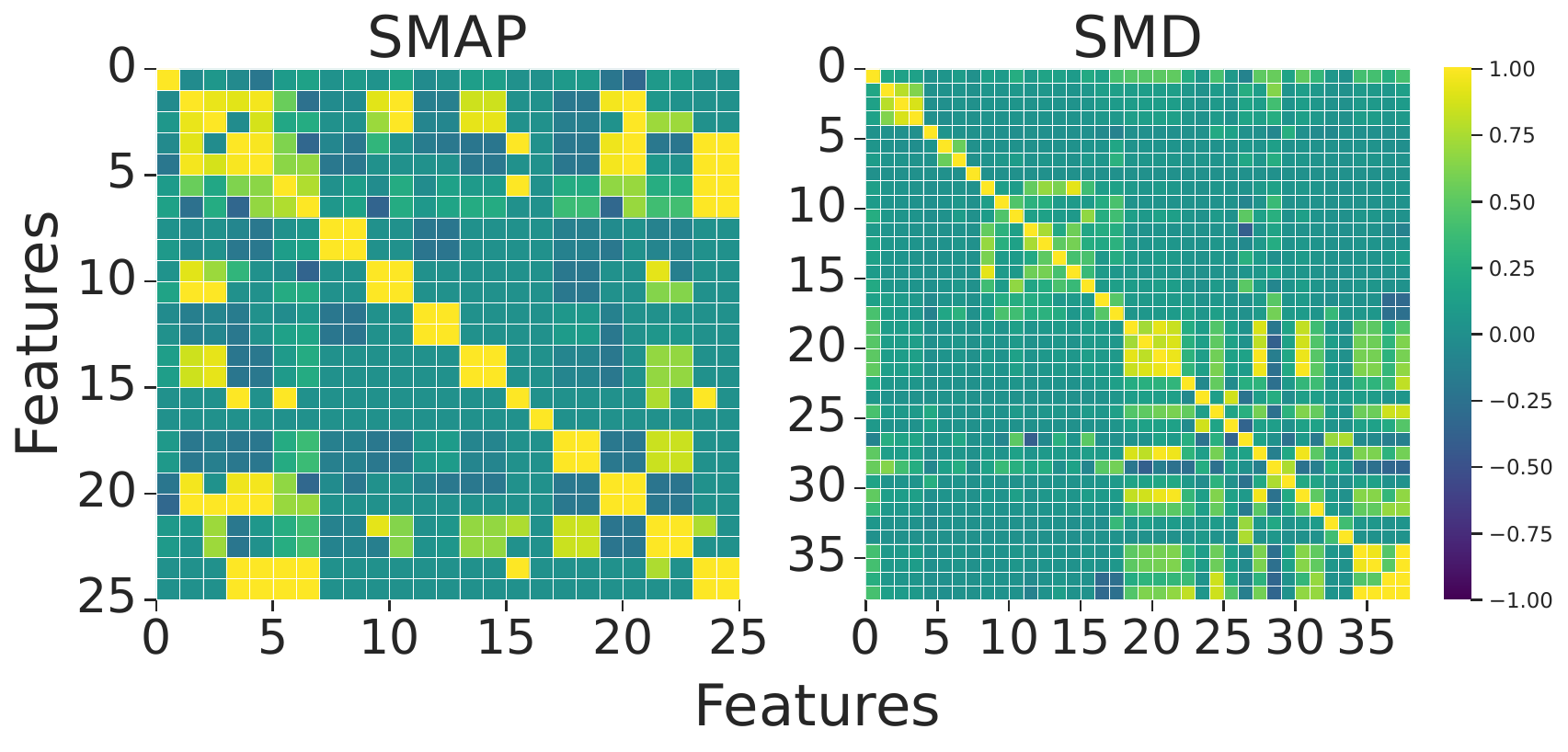}}
\vspace{-1em}
\caption{Visualization of correlation between each feature. It is closer to a positive (resp. negative) correlation when the color is brighter (resp. darker).}
\vspace{-2em}
\label{fig:correlation}
\end{wrapfigure}
As mentioned in the proposed method, the graph structure provides dependencies between features in the prediction process. As shown in Figure~\ref{fig:correlation}, anomaly detection benchmark datasets have a correlation between features. Therefore, the graph structure allows our model to capture correlations for each feature that might not be considered by separate training, which leads to a decrease in overall error for features. Figure~\ref{fig:Ablation} shows that our forecasting model with the graph structure showed lower errors than the model without graph structure in both continuous and discrete features. In other words, it is evident that the graph structure, considering the relationship between each feature, is imperative in anomaly detection datasets with correlations. 

\subsubsection{Separate Training of Continuous and Discrete Features.}
\begin{wrapfigure}{r}{8cm}
\vspace{-1em}
\centering
{\includegraphics[width=0.5\columnwidth]{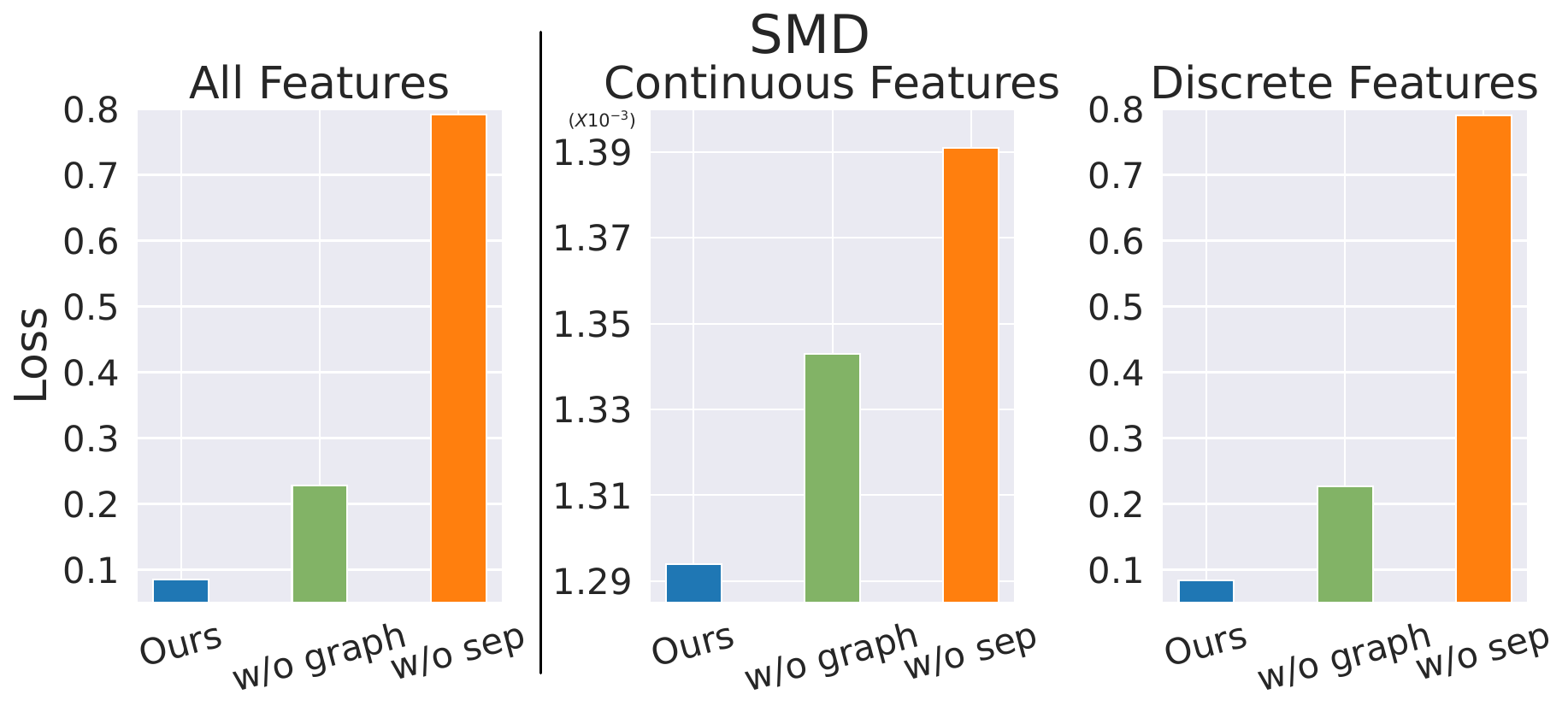}}
\caption{Ablation study results for graph structure and separate training of continuous (MSE loss) and discrete (CE loss) features.}
\label{fig:Ablation}
\vspace{-1em}
\end{wrapfigure}
We split the two groups of features to predict more accurately the time-series datasets for anomaly detection, which include both continuous and discrete features. In this subsection, we investigated the model without separation training to confirm the effectiveness of separation training. In order to implement the model without separation training, discrete features which consist of one-hot vectors (denoted $\mathbf{O}^{t}$) and continuous features were concatenated to be used as input to the continuous prediction model using a graph structure. In addition, it was trained with the MSE loss.

As a result, Figure~\ref{fig:Ablation} shows that the model without separation training significantly increases the cross-entropy loss compared to our model. In other words, it shows that separation training is necessary for datasets with continuous and discrete features, particularly in multivariate time-series anomaly detection.

\end{document}

%% file: math_commands.tex
\usepackage{amsmath,amsfonts,bm}

\def\eqref#1{equation~\ref{#1}}

\def\1{\bm{1}}

\DeclareMathAlphabet{\mathsfit}{\encodingdefault}{\sfdefault}{m}{sl}
\SetMathAlphabet{\mathsfit}{bold}{\encodingdefault}{\sfdefault}{bx}{n}













%% file: iclr2025.bbl
\begin{thebibliography}{26}
\providecommand{\natexlab}[1]{#1}
\providecommand{\url}[1]{\texttt{#1}}
\expandafter\ifx\csname urlstyle\endcsname\relax
  \providecommand{\doi}[1]{doi: #1}\else
  \providecommand{\doi}{doi: \begingroup \urlstyle{rm}\Url}\fi

\bibitem[Abdulaal et~al.(2021)Abdulaal, Liu, and Lancewicki]{abdulaal2021practical}
Ahmed Abdulaal, Zhuanghua Liu, and Tomer Lancewicki.
\newblock Practical approach to asynchronous multivariate time series anomaly detection and localization.
\newblock In \emph{Proceedings of the 27th ACM SIGKDD conference on knowledge discovery \& data mining}, pp.\  2485--2494, 2021.

\bibitem[Chen et~al.(2021)Chen, Chen, Zhang, Yuan, and Cheng]{chen2021learning}
Zekai Chen, Dingshuo Chen, Xiao Zhang, Zixuan Yuan, and Xiuzhen Cheng.
\newblock Learning graph structures with transformer for multivariate time-series anomaly detection in iot.
\newblock \emph{IEEE Internet of Things Journal}, 9\penalty0 (12):\penalty0 9179--9189, 2021.

\bibitem[Constantinescu \& Popescu(2023)Constantinescu and Popescu]{constantinescu2023interpolation}
Vlad-Raul Constantinescu and Ionel Popescu.
\newblock Interpolation property of shallow neural networks.
\newblock \emph{arXiv preprint arXiv:2304.10552}, 2023.

\bibitem[Deng \& Hooi(2021)Deng and Hooi]{deng2021graph}
Ailin Deng and Bryan Hooi.
\newblock Graph neural network-based anomaly detection in multivariate time series.
\newblock In \emph{Proceedings of the AAAI conference on artificial intelligence}, pp.\  4027--4035, 2021.

\bibitem[Garg et~al.(2022)Garg, Zhang, Samaran, Savitha, and Foo]{astha22evaluation}
Astha Garg, Wenyu Zhang, Jules Samaran, Ramasamy Savitha, and Chuan-Sheng Foo.
\newblock An evaluation of anomaly detection and diagnosis in multivariate time series.
\newblock \emph{IEEE Transactions on Neural Networks and Learning Systems}, 33\penalty0 (6):\penalty0 2508--2517, 2022.
\newblock \doi{10.1109/TNNLS.2021.3105827}.

\bibitem[Harvey et~al.(2022)Harvey, Naderiparizi, Masrani, Weilbach, and Wood]{harvey2022flexible}
William Harvey, Saeid Naderiparizi, Vaden Masrani, Christian Weilbach, and Frank Wood.
\newblock Flexible diffusion modeling of long videos.
\newblock \emph{Advances in Neural Information Processing Systems}, 35:\penalty0 27953--27965, 2022.

\bibitem[He \& Zhao(2019)He and Zhao]{he2019temporal}
Yangdong He and Jiabao Zhao.
\newblock Temporal convolutional networks for anomaly detection in time series.
\newblock In \emph{Journal of Physics: Conference Series}, pp.\  042050. IOP Publishing, 2019.

\bibitem[Hundman et~al.(2018)Hundman, Constantinou, Laporte, Colwell, and Soderstrom]{hundman2018detecting}
Kyle Hundman, Valentino Constantinou, Christopher Laporte, Ian Colwell, and Tom Soderstrom.
\newblock Detecting spacecraft anomalies using lstms and nonparametric dynamic thresholding.
\newblock In \emph{Proceedings of the 24th ACM SIGKDD international conference on knowledge discovery \& data mining}, pp.\  387--395, 2018.

\bibitem[Kim et~al.(2022)Kim, Choi, Choi, Lee, and Yoon]{kim2022towards}
Siwon Kim, Kukjin Choi, Hyun-Soo Choi, Byunghan Lee, and Sungroh Yoon.
\newblock Towards a rigorous evaluation of time-series anomaly detection.
\newblock In \emph{Proceedings of the AAAI Conference on Artificial Intelligence}, pp.\  7194--7201, 2022.

\bibitem[Kingma \& Ba(2014)Kingma and Ba]{kingma2014adam}
Diederik~P Kingma and Jimmy Ba.
\newblock Adam: A method for stochastic optimization.
\newblock \emph{arXiv preprint arXiv:1412.6980}, 2014.

\bibitem[Li et~al.(2022)Li, Zhao, Hu, Botta, Ionescu, and Chen]{li2022ecod}
Zheng Li, Yue Zhao, Xiyang Hu, Nicola Botta, Cezar Ionescu, and George Chen.
\newblock {ECOD}: Unsupervised outlier detection using empirical cumulative distribution functions.
\newblock \emph{IEEE Transactions on Knowledge and Data Engineering}, 2022.

\bibitem[Malhotra et~al.(2015)Malhotra, Vig, Shroff, Agarwal, et~al.]{malhotra2015long}
Pankaj Malhotra, Lovekesh Vig, Gautam Shroff, Puneet Agarwal, et~al.
\newblock Long short term memory networks for anomaly detection in time series.
\newblock In \emph{Esann}, volume 2015, pp.\ ~89, 2015.

\bibitem[Munir et~al.(2018)Munir, Siddiqui, Dengel, and Ahmed]{munir2018deepant}
Mohsin Munir, Shoaib~Ahmed Siddiqui, Andreas Dengel, and Sheraz Ahmed.
\newblock Deepant: A deep learning approach for unsupervised anomaly detection in time series.
\newblock \emph{{IEEE} Access}, 7:\penalty0 1991--2005, 2018.

\bibitem[Nie et~al.(2022)Nie, Nguyen, Sinthong, and Kalagnanam]{nie2022time}
Yuqi Nie, Nam~H Nguyen, Phanwadee Sinthong, and Jayant Kalagnanam.
\newblock A time series is worth 64 words: Long-term forecasting with transformers.
\newblock \emph{arXiv preprint arXiv:2211.14730}, 2022.

\bibitem[Raissi et~al.(2019)Raissi, Perdikaris, and Karniadakis]{raissi2019physics}
Maziar Raissi, Paris Perdikaris, and George~E Karniadakis.
\newblock Physics-informed neural networks: A deep learning framework for solving forward and inverse problems involving nonlinear partial differential equations.
\newblock \emph{Journal of Computational physics}, 378:\penalty0 686--707, 2019.

\bibitem[Ruff et~al.(2018)Ruff, Vandermeulen, Goernitz, Deecke, Siddiqui, Binder, M{\"u}ller, and Kloft]{ruff2018deep}
Lukas Ruff, Robert Vandermeulen, Nico Goernitz, Lucas Deecke, Shoaib~Ahmed Siddiqui, Alexander Binder, Emmanuel M{\"u}ller, and Marius Kloft.
\newblock Deep one-class classification.
\newblock In \emph{International conference on machine learning}, pp.\  4393--4402. PMLR, 2018.

\bibitem[Sch{\"o}lkopf et~al.(1999)Sch{\"o}lkopf, Williamson, Smola, Shawe-Taylor, and Platt]{scholkopf1999support}
Bernhard Sch{\"o}lkopf, Robert~C Williamson, Alex Smola, John Shawe-Taylor, and John Platt.
\newblock Support vector method for novelty detection.
\newblock \emph{Advances in neural information processing systems}, 12, 1999.

\bibitem[Su et~al.(2019)Su, Zhao, Niu, Liu, Sun, and Pei]{su2019robust}
Ya~Su, Youjian Zhao, Chenhao Niu, Rong Liu, Wei Sun, and Dan Pei.
\newblock Robust anomaly detection for multivariate time series through stochastic recurrent neural network.
\newblock In \emph{Proceedings of the 25th ACM SIGKDD international conference on knowledge discovery \& data mining}, pp.\  2828--2837, 2019.

\bibitem[Tashiro et~al.(2021)Tashiro, Song, Song, and Ermon]{tashiro2021csdi}
Yusuke Tashiro, Jiaming Song, Yang Song, and Stefano Ermon.
\newblock Csdi: Conditional score-based diffusion models for probabilistic time series imputation.
\newblock \emph{Advances in Neural Information Processing Systems}, 34:\penalty0 24804--24816, 2021.

\bibitem[Wagner et~al.(2023)Wagner, Michels, Schulz, Nair, Rudolph, and Kloft]{Wagner23timesead}
Dennis Wagner, Tobias Michels, Florian C.~F. Schulz, Arjun Nair, Maja Rudolph, and Marius Kloft.
\newblock {TimeSeAD}: Benchmarking deep multivariate time-series anomaly detection.
\newblock \emph{Trans. Mach. Learn. Res.}, 2023, 2023.
\newblock URL \url{https://openreview.net/forum?id=iMmsCI0JsS}.

\bibitem[Wu et~al.(2022)Wu, Hu, Liu, Zhou, Wang, and Long]{wu2022timesnet}
Haixu Wu, Tengge Hu, Yong Liu, Hang Zhou, Jianmin Wang, and Mingsheng Long.
\newblock Timesnet: Temporal 2d-variation modeling for general time series analysis.
\newblock \emph{arXiv preprint arXiv:2210.02186}, 2022.

\bibitem[Xu et~al.(2018)Xu, Chen, Zhao, Li, Bu, Li, Liu, Zhao, Pei, Feng, et~al.]{xu2018unsupervised}
Haowen Xu, Wenxiao Chen, Nengwen Zhao, Zeyan Li, Jiahao Bu, Zhihan Li, Ying Liu, Youjian Zhao, Dan Pei, Yang Feng, et~al.
\newblock Unsupervised anomaly detection via variational auto-encoder for seasonal kpis in web applications.
\newblock In \emph{Proceedings of the 2018 world wide web conference}, pp.\  187--196, 2018.

\bibitem[Yousefzadeh(2021)]{yousefzadeh2021deep}
Roozbeh Yousefzadeh.
\newblock Deep learning generalization and the convex hull of training sets.
\newblock \emph{arXiv preprint arXiv:2101.09849}, 2021.

\bibitem[Yousefzadeh \& Huang(2020)Yousefzadeh and Huang]{yousefzadeh2020using}
Roozbeh Yousefzadeh and Furong Huang.
\newblock Using wavelets and spectral methods to study patterns in image-classification datasets.
\newblock \emph{arXiv preprint arXiv:2006.09879}, 2020.

\bibitem[Zeng et~al.(2023)Zeng, Chen, Zhang, and Xu]{zeng2023transformers}
Ailing Zeng, Muxi Chen, Lei Zhang, and Qiang Xu.
\newblock Are transformers effective for time series forecasting?
\newblock In \emph{Proceedings of the AAAI conference on artificial intelligence}, pp.\  11121--11128, 2023.

\bibitem[Zhao et~al.(2020)Zhao, Wang, Duan, Huang, Cao, Tong, Xu, Bai, Tong, and Zhang]{zhao2020multivariate}
Hang Zhao, Yujing Wang, Juanyong Duan, Congrui Huang, Defu Cao, Yunhai Tong, Bixiong Xu, Jing Bai, Jie Tong, and Qi~Zhang.
\newblock Multivariate time-series anomaly detection via graph attention network.
\newblock In \emph{2020 IEEE International Conference on Data Mining (ICDM)}, pp.\  841--850. IEEE, 2020.

\end{thebibliography}
